\documentclass[preprint,5p,authoryear]{elsarticle}

\usepackage{amsmath,amssymb,amsfonts}
\usepackage{algpseudocode,algorithm}
\usepackage{subfigure}
\usepackage{graphicx}

\journal{Neural Networks}

\begin{document}

\sloppy

\begin{frontmatter}



\title{Optimistic Reinforcement Learning by\\Forward Kullback-Leibler Divergence Optimization}


\author{Taisuke Kobayashi\corref{cor}}
\ead{kobayashi@is.naist.jp}
\ead[url]{http://kbys\_t.gitlab.io/en/}

\cortext[cor]{Corresponding author}

\address{Nara Institute of Science and Technology, Nara, Japan}

\begin{abstract}

This paper addresses a new interpretation of the traditional optimization method in reinforcement learning (RL) as optimization problems using reverse Kullback-Leibler (KL) divergence, and derives a new optimization method using forward KL divergence, instead of reverse KL divergence in the optimization problems.
Although RL originally aims to maximize return indirectly through optimization of policy, the recent work by Levine has proposed a different derivation process with explicit consideration of optimality as stochastic variable.
This paper follows this concept and formulates the traditional learning laws for both value function and policy as the optimization problems with reverse KL divergence including optimality.
Focusing on the asymmetry of KL divergence, the new optimization problems with forward KL divergence are derived.
Remarkably, such new optimization problems can be regarded as optimistic RL.
That optimism is intuitively specified by a hyperparameter converted from an uncertainty parameter.
In addition, it can be enhanced when it is integrated with prioritized experience replay and eligibility traces, both of which accelerate learning.
The effects of this expected optimism was investigated through learning tendencies on numerical simulations using Pybullet.
As a result, moderate optimism accelerated learning and yielded higher rewards.
In a realistic robotic simulation, the proposed method with the moderate optimism outperformed one of the state-of-the-art RL method.

\end{abstract}

\begin{keyword}



Reinforcement learning \sep Control as probabilistic inference \sep Kullback-Leibler divergence \sep Optimistic learning

\end{keyword}

\end{frontmatter}


\section{Introduction}
\label{sec:introduction}

Reinforcement learning (RL)~\citep{sutton2018reinforcement} is one of the promising approaches to acquire the optimal policy for controlling complicated systems like robotic manipulation/locomotion~\citep{levine2018learning,peng2017deeploco}.
In the last decade, a lot of practical applications have been reported~\citep{modares2015optimized,tsurumine2019deep} due to remarkable improvements in function approximation performance by deep neural networks~\citep{lecun2015deep}, which also removed limitations on state input.
In practice, however, RL is still not enough reliable and requires great efforts for trial and error.

Recent studies towards improving the performance of RL, especially in terms of learning stability and sample efficiency, have been central topics.
For example, in terms of learning stability, frameworks for
reducing the estimation bias of value function~\citep{schulman2015high,fujimoto2018addressing,vuong2019uncertainty,kobayashi2021t}
and regularization techniques for policy updates~\citep{schulman2015trust,munos2016safe,haarnoja2018soft,parisi2019td,schulman2017proximal}
have been widely proposed, and they have succeeded in reducing the risk of overlearning and trapping into local solutions.
In terms of sample efficiency, reuse of experienced data is effective,
and direct reuse by experience replay~\citep{lin1992self,schaul2015prioritized,andrychowicz2017hindsight}
and model-based RL using learned state transition model~\citep{vuong2019uncertainty,chua2018deep,clavera2019model}
have shown remarkable results.

However, these improvements have been accomplished by addon-like extensions to the underlying fundamental RL learning law~\citep{sutton2018reinforcement}.
Although steady improvements can be expected, it is difficult to fully eliminate the inherent problems of the current RL.
Therefore, this study is motivated by the need to discover and investigate new learning laws of the optimal policy by interpreting the traditional RL learning law from different perspectives.

In this direction, \textit{RL as probabilistic inference}, as proposed by \citet{levine2018reinforcement}, stands on a very important position.
It allows existing optimal control problems including RL to be interpreted as inference problems by explicitly giving the optimality as binary stochastic variable.
For example, a soft actor critic (SAC) algorithm has been re-derived in the original paper, and new learning algorithms, such as policy optimization with stochastic latent space~\citep{lee2020stochastic} and variational-inference-based model predictive control~\citep{okada2020variational}, have also been proposed using this concept.

With this concept, this paper reveals that the optimizations of both value function and policy on the traditional RL are consistent with \textit{reverse} Kullback-Leibler (KL) divergence optimization (named RKL-RL in this paper).
Specifically, the update law of the traditional value function can be derived by considering the problem of minimizing reverse KL divergence between the probabilities of optimality with different conditions.
The traditional policy improvement can also be derived by minimizing/maximizing reverse KL divergences with optimal/non-optimal policies inferred from Bayesian theorem, respectively.
That is, a new interpretation, \textit{RL as divergence optimization}, is suggested.
According to this suggestion, a new optimization method, so-called FKL-RL, is formulated based on \textit{forward} KL divergence optimization, which replaces the reverse KL divergences in the above optimization problems with the corresponding forward KL divergences.
Note that, only regarding the policy optimization, it is pointed out that traditional RL uses reverse KL divergence, while imitation learning uses forward KL divergence~\citep{ke2019imitation,uchibe2020imitation}.

These two types of optimization problems yield similar but different learning laws.
One remarkable difference is optimism in FKL-RL.
The optimism in the traditional RL has been implemented by several ways:
such as the optimistic initialization of value function~\citep{machado2015domain,rashid2019optimistic};
and the use of upper confidence~\citep{sunehag2015rationality,curi2020efficient}.
Indeed, biological decision making has also been suggested to be optimistic RL~\citep{lefebvre2017behavioural}, which has asymmetric learning rates.
Although these are given based on the traditional RL with several modifications, FKL-RL naturally possesses that property.
This optimism can be controlled by a heuristic design of an uncertainty parameter left only in FKL-RL.
In addition, FKL-RL integrated with prioritized experience replay~\citep{schaul2015prioritized} and/or eligibility traces~\citep{sutton2018reinforcement}, both of which are well-known to accelerate learning, would make the optimism enhance.

To investigate the effects of this expected optimism, numerical simulations using Pybullet~\citep{coumans2016pybullet} are conducted with RKL-RL and FKL-RL.
Their learning tendencies indicates that optimism gained by FKL-RL certainly enables the agent to escape from local optima, although too much optimism increases the variance of learning results, as in the case of policy gradient methods without a baseline term~\citep{greensmith2004variance}.
In a more practical robot simulation, FKL-RL is able to obtain performance better than one of the latest methods~\citep{schulman2017proximal}.

The contributions in this paper are three folds.
\begin{enumerate}
    \item The traditional optimization method in RL was transformed into the optimization problems using reverse KL divergence.
    \item FKL-RL was newly derived by considering another divergence (i.e. forward KL divergence) for the defined optimization problems.
    \item The biological optimism was found in FKL-RL as the remarkable difference from RKL-RL, and experimentally investigated through numerical simulations.
\end{enumerate}
RL as optimization problem with one of the divergences, as derived in this paper, may become a new step towards breakthrough in the traditional RL.

\section{Preliminaries}
\label{sec:preliminary}

\subsection{Traditional reinforcement learning}

The purpose of the traditional RL is to optimize policy so that the accumulation of rewards in the future (so-called return) is maximized~\citep{sutton2018reinforcement}.
This problem can be solved in Markov decision process (MDP) with the tuple $(\mathcal{S}, \mathcal{A}, \mathcal{R}, p_0, p_e)$.
Specifically, MDP assumes the following process.
The state $s \in \mathcal{S}$ is first sampled from environment with either of probabilities, $s \sim p_0(s)$ as the initial random state or $s^\prime \sim p_e(s^\prime \mid s, a)$ as the state transition.
According to $s$, an agent decides the action $a \in \mathcal{A}$, using the learnable policy $a \sim \pi(a \mid s)$.
$a$ acts on the environment and stochastically updates the state to the next one $s^\prime \sim p_e(s^\prime \mid s, a)$.
At that time, the agent obtains a reward $r \in \mathcal{R}$ from the environment: $r = r(s, a)$.
By repeating this process with the experienced data $(s, a, s^\prime, r)$, the agent obtains the following return $R_t$ from the current time step $t$.
\begin{align}
    R_t = \sum_{k=0}^\infty \gamma^k r_{t+k}
    \label{eq:return}
\end{align}
where $\gamma \in [0, 1)$ denotes the discount factor.
Note that, by multiplying $R_t$ with $(1 - \gamma)$, the scale of return can be normalized to the scale of reward while holding the optimization problem.

Under this MDP, the optimization problem of $\pi(a \mid s)$ is given as follows:
\begin{align}
    \pi^*(a \mid s) = \arg \max_{\pi} \mathbb{E}_{p_e, \pi}[R_t \mid s_t = s]
    \label{eq:prob_rl}
\end{align}
Here, the expected value of return is utilized since the future rewards are not deterministically given.

\subsubsection{Value function optimization by Bellman equation}\label{subsubsec:trad_value}

To solve the above optimization problem, state/action value functions are defined to estimate the expected return as $V(s) = \mathbb{E}_{p_e, \pi}[R_t \mid s_t = s]$ and $Q(s, a) = \mathbb{E}_{p_e, \pi}[R_t \mid s_t = s, a_t = a]$, respectively.
Note that these two have the relationship, $V(s) = \mathbb{E}_{\pi}[Q(s, a)]$.
If we know the current value, we can solve the above problem by simply taking actions that yield best (or better at least) values.

The value function can be learned according to Bellman equation, which focuses on the recursive relationship of return.
\begin{align}
    V(s) &= \mathbb{E}_{p_e, \pi}[r_t]
    \nonumber \\
    &+ \gamma \mathbb{E}_{p_e, \pi}[\sum_{k=0}^\infty \gamma^k r_{t+1+k} \mid s_{t+1} \sim p_e(s^\prime \mid s = s_t, a = a_t)]
    \nonumber \\
    &\simeq r_t + \gamma V(s^\prime = s_{t+1})
\end{align}
where the last approximation is given by one-step Monte Carlo method.
Since the value function holding this recursive equation is accurate, $V(s)$ can be learned by minimizing temporal difference (TD) error, $\delta(s, a) := r + \gamma V(s^\prime) - V(s)$ with $s^\prime \sim p_e(s^\prime \mid s, a)$, as follows:
\begin{align}
    V^*(s) = \arg \min_{V(s)} \cfrac{1}{2} \delta^2(s, a)
    \label{eq:prob_value}
\end{align}
In RL with deep neural networks, $r + \gamma V(s^\prime)$ is regarded as the target signal and generated from target network like \citet{kobayashi2021t} did.
Note that there are several variants such as Q-learning~\citep{sutton2018reinforcement}, but they are not discussed in this paper for brevity.

\subsubsection{Policy optimization by policy gradient}

Using the value function, the policy can directly be optimized via eq.~\eqref{eq:prob_rl}.
That is, the following maximization problem is solved by the gradients w.r.t. the policy.
\begin{align}
    \pi^*(a \mid s) &= \arg \max_{\pi} \mathbb{E}_{\pi}[Q(s, a) - V(s)]
    \nonumber \\
    \nabla \mathbb{E}_{\pi}[Q(s, a) - V(s)] &= \mathbb{E}_{\pi}[(Q(s, a) - V(s))\nabla \ln \pi(a \mid s)]
    \label{eq:prob_policy}
\end{align}
Note that $V(s)$ in the first line is added to reduce variance of learning~\citep{greensmith2004variance}, although its gradient is expected to be zero analytically.
This optimization form is well-known as the actor-critic method~\citep{sutton2018reinforcement}.

\subsection{Forward/reverse Kullback-Leibler divergence}

KL divergence is one of the famous divergence measures between two probability distributions~\citep{kullback1997information}.
Given $p(x)$ and $q(x)$ as the probabilities for stochastic variable $x$, KL divergence is defined as follows:
\begin{align}
    \mathrm{KL}(p(x) \mid q(x)) := \int_x p(x) \ln \cfrac{p(x)}{q(x)} dx
    \label{eq:def_kl}
\end{align}
where the integral operation is given only when $p$ and $q$ are continuous (i.e. probability density functions), and switched to the summation operation when they are discrete (i.e. probability mass functions).

As can be seen in the above definition, KL divergence is asymmetry: $\mathrm{KL}(p(x) \mid q(x)) \neq \mathrm{KL}(q(x) \mid p(x))$.
Therefore, in particular when considering optimization problems with KL divergence, we often distinguish forward or reverse KL divergence by which a target, $p(x)$, and a model to be optimized, $q(x)$, are entered into left or right side.
\begin{align}
    \begin{cases}
        \mathrm{KL}(p(x) \mid q(x)) & \mathrm{Forward}
        \\
        \mathrm{KL}(q(x) \mid p(x)) & \mathrm{Reverse}
    \end{cases}
    \label{eq:def_rfkl}
\end{align}

\section{Proposal}
\label{sec:proposal}

\subsection{Introduction of optimality}

\citet{levine2018reinforcement} has proposed a new interpretation of RL as probabilistic inference.
With reference to the paper, an optimal variable $O = \{0, 1\}$ is introduced.
For simplicity, this paper supposes that $O$ represents the optimality from the current state to the future.
That is, probability mass functions, $p(O=1 \mid s) \in (0, 1)$ and $p(O=1 \mid s, a) \in (0, 1)$, can be assumed as the functions of $V(s)$ and $Q(s, a)$, respectively.
By definition, arbitrary monotonic function (but satisfying $(0, 1)$) is acceptable, but in this study, following previous studies like \citet{levine2018reinforcement}, the following exponential function with an uncertainty parameter $\tau > 0$ is employed as an example.
\begin{align}
    p(O=1 \mid s) &= \exp \left ( \cfrac{V(s) - C}{\tau} \right ) =: p_V(s)
    \label{eq:pv} \\
    p(O=1 \mid s, a) &= \exp \left ( \cfrac{Q(s, a) - C}{\tau} \right ) =: p_Q(s, a)
    \label{eq:pq}
\end{align}
where $C$ is a constant to satisfy $V(s) - C \leq 0$ and $Q(s, a) - C \leq 0$ for $p(O) \in (0, 1)$.
Note that the boundary (i.e. $0$ and $1$) is excluded since it is infeasible to make the current situation perfectly optimal or non-optimal.
With this natural assumption, all the following operations can be performed without exception.
Since $O$ is binary, the probability of $O=0$ can be given as $p(O=0 \mid s) = 1 - p_V(s)$ and $p(O=0 \mid s, a) = 1 - p_Q(s, a)$.

Using these probabilities, the optimal and non-optimal policies are defined.
Specifically, Bayesian theorem for $\pi(a \mid s, O)$ derives the following relationship with the baseline policy $b(a \mid s)$, which is used for sampling actions.
\begin{align}
    \pi(a \mid s, O) &= \cfrac{p(O \mid s, a) b(a \mid s)}{p(O \mid s)}
    \nonumber \\
    &= \begin{cases}
        \cfrac{p_Q(s, a)}{p_V(s)} b(a \mid s) =: \pi^+(a \mid s) & O = 1
        \\
        \cfrac{1 - p_Q(s, a)}{1 - p_V(s)} b(a \mid s) =: \pi^-(a \mid s)& O = 0
    \end{cases}
    \label{eq:policy_opt}
\end{align}

\subsection{Minimization of KL divergences between two optimal probabilities}

Here, the optimization of the value function $V(s)$ is considered.
Note that if $Q(s, a)$ is desired to be optimized directly, we may use the same derivation procedure combined with the fact $V(s) = \mathbb{E}_\pi[Q(s, a)]$.
In this paper, however, only $V(s)$ is optimized with the approximation $Q(s, a) \simeq r + \gamma V(s^\prime)$ as the target signal, like the traditional RL (see the section~\ref{subsubsec:trad_value}).

It is essential to back-propagate the information increased by the addition of action $a$ (and reward $r$ returned as a result of $a$) as a condition to the probability where $a$ is not included in the condition.
This can be achieved by updating $p(O \mid s)$, which is the function of $V$, so as to minimize divergence between $p(O \mid s, a)$ and $p(O \mid s)$.
As representative measure of divergence, this paper employs forward/reverse KL divergences, as mentioned in the introduction.

\subsubsection{Reverse KL divergence}

First, we consider the following minimization problem with reverse KL divergence.
\begin{align}
    \min \mathbb{E}_{p_e, b}[\mathrm{KL}(p(O \mid s) \mid p(O \mid s, a))]
    \label{eq:prob_kl_rv}
\end{align}
When $V(s)$ is parameterized by $\theta$, the gradient of the reverse KL divergence, which is inside the expectation, over $\theta$, $g_\theta^R$, is derived as follows:
\begin{align}
    g_\theta^R &=
    \nabla_\theta p_V(s) \ln \cfrac{p_V(s)}{p_Q(s, a)}
    - \nabla_\theta p_V(s) \ln \cfrac{1 - p_V(s)}{1 - p_Q(s, a)}
    \nonumber \\
    &+ p_V(s) \nabla_\theta \ln p_V(s)
    + (1 - p_V(s)) \nabla_\theta \ln (1 - p_V(s))
    \nonumber \\
    &= - g_\theta^V \cfrac{p_V(s)}{\tau}
    \left \{ \cfrac{Q(s, a) - V(s)}{\tau}
    + \ln \cfrac{1 - p_V(s)}{1 - p_Q(s, a)} \right \}
    \nonumber \\
    &\propto - g_\theta^V
    \left \{ (Q(s, a) - V(s))
    + \tau \ln \cfrac{1 - p_V(s)}{1 - p_Q(s, a)} \right \}
    \label{eq:grad_kl_rv_origin}
\end{align}
where $g_\theta^V = \nabla_\theta V(s)$ for simplicity.
The two terms in the second line cancel each other out.
The last proportion is given by multiplying the third line by $\tau^2 p_V(s)^{-1}$.
Such a multiplication for analytical computation will also be introduced later, but note that it would induce a bias, although the update direction does not change.
That is, almost the gradients derived in this paper are biased but consistent with the strict ones.

After this, the gradients can be computed analytically by multiplying them in the same way, but note that this can introduce a strict bias, although the update direction itself does not change.

Unfortunately, the above gradient cannot be computed due to the existence of $C$ inside $p_V(s)$ and $p_Q(s, a)$.
$C$ is defined as the constant, but in general, its value is task-specific and unknown.
Therefore, by considering the limit of $\tau \to 0$, $g_\theta^R$ can be approximated in a computable way.
\begin{align}
    g_\theta^R \simeq - g_\theta^V (Q(s, a) - V(s))
    \label{eq:grad_kl_rv_approx}
\end{align}
As a remark, this gradient is equivalent with one for the squared error shown in eq.~\eqref{eq:prob_value}.
In other words, the update law of the value function in the traditional RL can be regarded as the minimization problem of the reverse KL divergence.

\subsubsection{Forward KL divergence}

Next, we consider the following minimization problem with forward KL divergence.
\begin{align}
    \min \mathbb{E}_{p_e, b}[\mathrm{KL}(p(O \mid s, a) \mid p(O \mid s))]
    \label{eq:prob_fkl}
\end{align}
Similar to the above procedure, the gradient of the forward KL divergence over $\theta$, $g_\theta^F$, is derived as follows:
\begin{align}
    g_\theta^F &=
    - p_Q(s, a) \nabla_\theta \! \ln p_V(s)
    \!-\! (1 - p_Q(s, a)) \nabla_\theta \! \ln (1 - p_V(s))
    \nonumber \\
    &= - \cfrac{g_\theta^V}{\tau}
    \left \{ p_Q(s, a) - \cfrac{p_V(s) (1 - p_Q(s, a))}{1 - p_V(s)} \right \}
    \nonumber \\
    &= - \cfrac{g_\theta^V}{\tau}
    \cfrac{p_Q(s, a) - p_V(s)}{1 - p_V(s)}
    \nonumber \\
    &\propto - g_\theta^V
    \tau \left ( \cfrac{p_Q(s, a)}{p_V(s)} - 1 \right )
    \nonumber \\
    &= - g_\theta^V
    \tau \left \{ \exp \left (\cfrac{Q(s, a) - V(s)}{\tau} \right ) - 1 \right \}
    \label{eq:grad_kl_fv}
\end{align}
where the proportion is given by multiplying the third line by $\tau^2 (1 - p_V(s)) p_V(s)^{-1}$.
Unlike eq.~\eqref{eq:grad_kl_rv_origin}, the above equation can directly be computed since $C$ is excluded in the derivation process.
The difference with eq.~\eqref{eq:grad_kl_rv_approx} is discussed later in the section~\ref{sec:analysis}.

\subsection{Optimization of KL divergences from optimal/non-optimal policies}
\label{subsec:opt_policy}

\begin{figure}[tb]
    \centering
    \subfigure[Before optimization]{
        \includegraphics[keepaspectratio=true,width=0.95\linewidth]{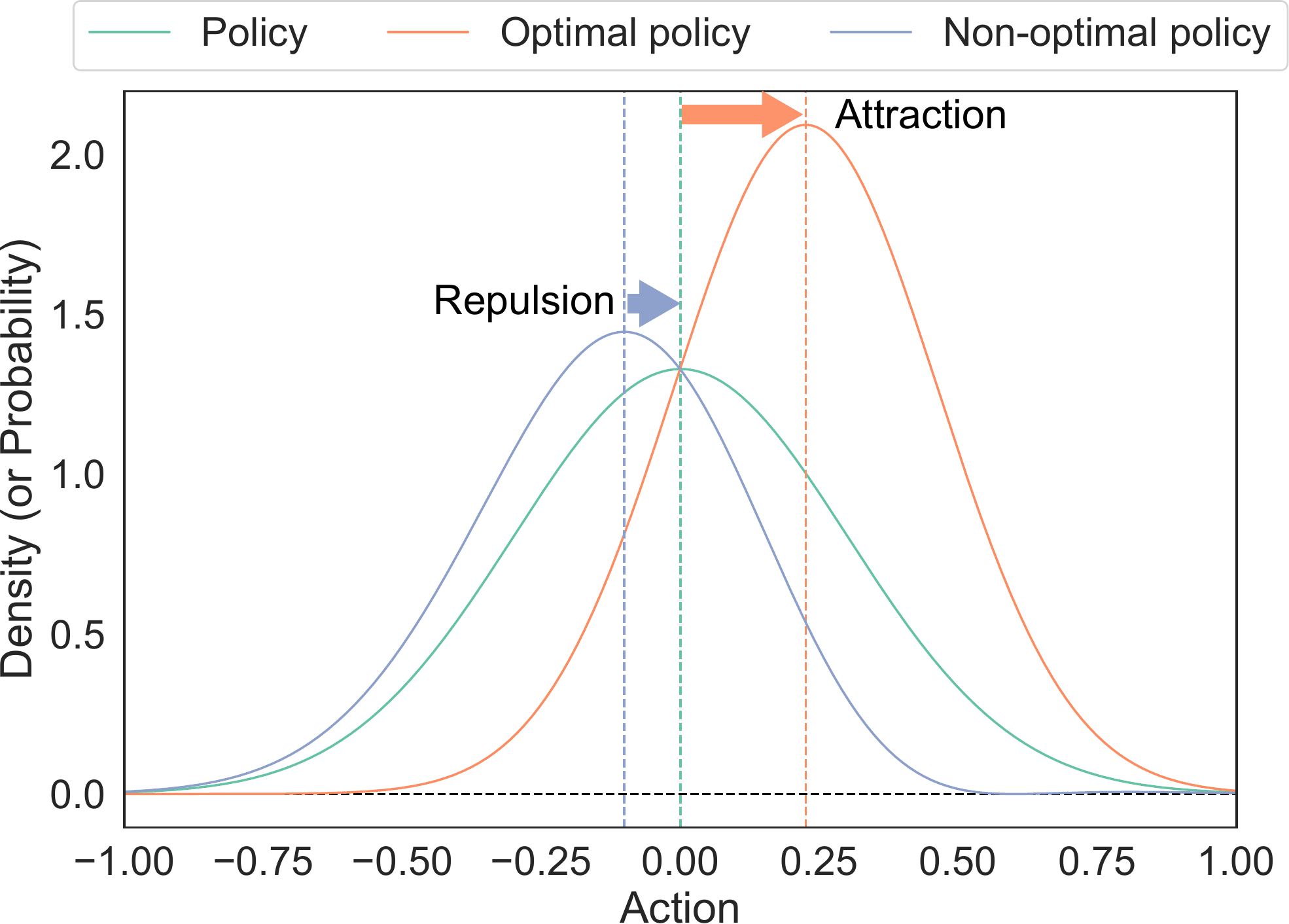}
    }
    \centering
    \subfigure[After optimization]{
        \includegraphics[keepaspectratio=true,width=0.95\linewidth]{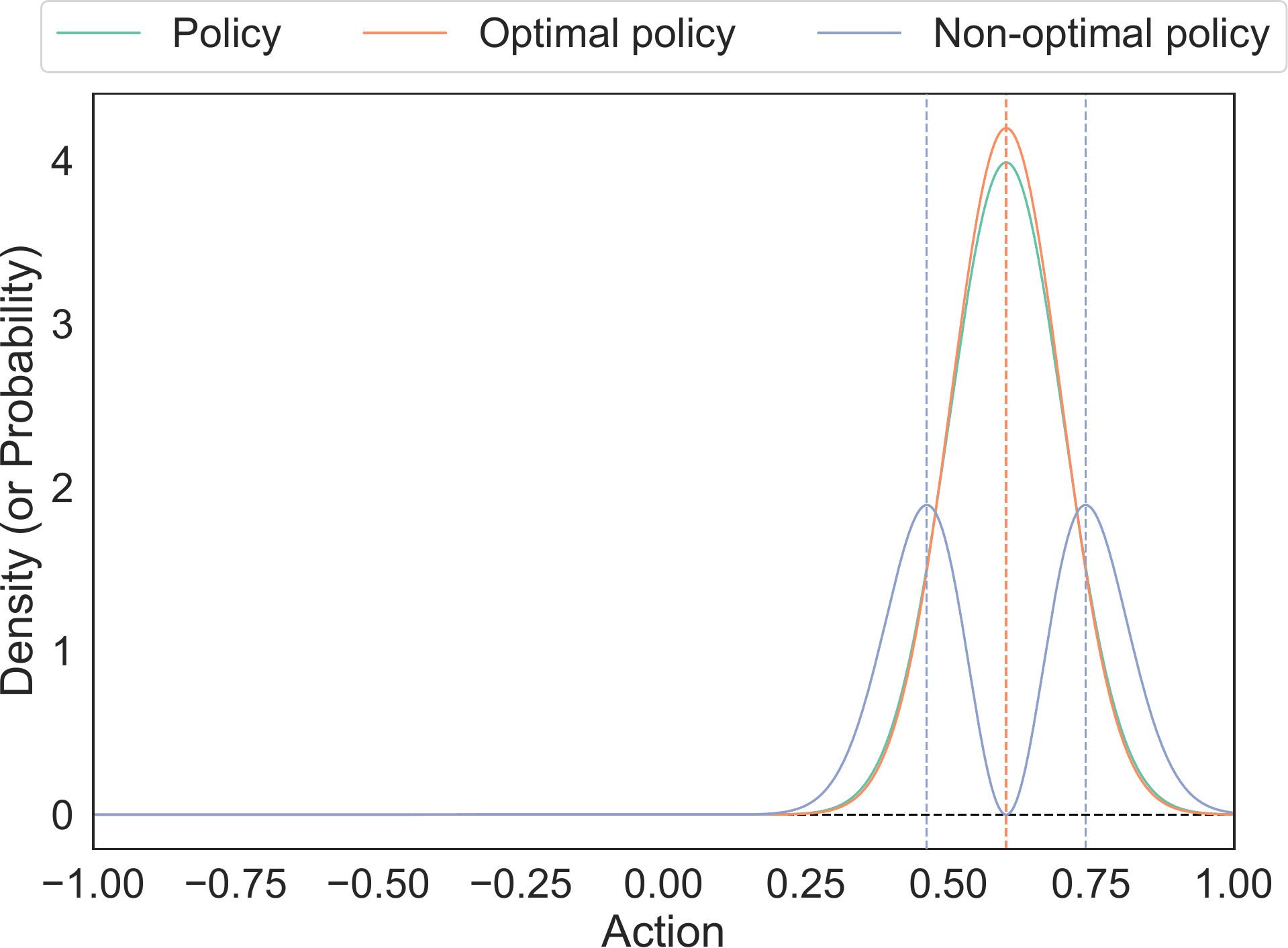}
    }
    \caption{Illustration of optimization problem of policy inspired by triplet loss:
        (a) the learnable policy is attracted to the optimal policy and repulsed by the non-optimal policy;
        (b) finally, the learnable policy almost converges to the optimal policy, and the non-optimal policy keeps the learnable policy in place.
    }
    \label{fig:img_triplet}
\end{figure}

From here, the optimization of the policy $\pi(a \mid s)$ with parameter set $\phi$ is considered.
The optimal and non-optimal policies ($\pi^+(a \mid s)$ and $\pi^-(a \mid s)$, respectively) are already given in eq.~\eqref{eq:policy_opt}.
Using them, an optimization problem is raised such that $\pi(a \mid s)$ approaches $\pi^+(a \mid s)$ and moves away from $\pi^-(a \mid s)$ (see Fig.~\ref{fig:img_triplet}), inspired by triplet loss~\citep{schultz2003learning,chechik2010large}.
Since we still need to evaluate the divergence between policies as in the case of $V(s)$, the forward/reverse KL divergences are employed again in this paper.

\subsubsection{Reverse KL divergence}

First, the following minimization problem with reverse KL divergence is solved.
\begin{align}
    \min \mathbb{E}_{p_e}[\mathrm{KL}(\pi(a \mid s) \mid \pi^+(a \mid s)) - \mathrm{KL}(\pi(a \mid s) \mid \pi^-(a \mid s))]
    \label{eq:prob_kl_rp}
\end{align}
The gradient of the two terms inside the expectation over $\phi$, $g_\phi^R$, is derived as follows:
\begin{align}
    g_\phi^R &=
    \int_a
    \nabla_\phi \pi(a \mid s) \ln \cfrac{\pi(a \mid s)}{\pi^+(a \mid s)}
    - \nabla_\phi \pi(a \mid s) \ln \cfrac{\pi(a \mid s)}{\pi^-(a \mid s)}
    \nonumber \\
    &+ \pi(a \mid s) \nabla_\phi \ln \pi(a \mid s)
    - \pi(a \mid s) \nabla_\phi \ln \pi(a \mid s)
    da
    \nonumber \\
    &= \int_a
    \pi(a \mid s) \cfrac{\nabla_\phi \pi(a \mid s)}{\pi(a \mid s)}
    \ln \cfrac{\pi^-(a \mid s)}{\pi^+(a \mid s)} da
    \nonumber \\
    &= \mathbb{E}_{\pi} \left [ g_\phi^\pi
    \left \{ \cfrac{V(s) - Q(s, a)}{\tau}
    + \ln \cfrac{1 - p_Q(s, a)}{1 - p_V(s)} \right \} \right ]
    \nonumber \\
    &\propto \mathbb{E}_{\pi} \left [ - g_\phi^\pi
    \left \{ (Q(s, a) - V(s))
    + \tau \ln \cfrac{1 - p_V(s)}{1 - p_Q(s, a)} \right \} \right ]
    \label{eq:grad_kl_rp_origin}
\end{align}
where $g_\phi^\pi = \nabla_\phi \ln \pi(a \mid s)$.
The last proportion is given by multiplying the third line by $\tau$.

We can see that the same non-computable term as in eq.~\eqref{eq:grad_kl_rv_origin} appears.
In the same way, i.e. by considering the limit of $\tau \to 0$, it can be excluded.
\begin{align}
    g_\phi^R &\simeq \mathbb{E}_{\pi} \left [ - g_\phi^\pi
    (Q(s, a) - V(s)) \right ]
    \nonumber \\
    &= \mathbb{E}_{b} \left [- \cfrac{\pi(a \mid s)}{b(a \mid s)} g_\phi^\pi
    (Q(s, a) - V(s)) \right ]
    \label{eq:grad_kl_rp_approx}
\end{align}
where importance sampling~\citep{tokdar2010importance} is applied to sample $a$ from the baseline policy $b(a \mid s)$.
As a result, the minimization problem given in eq.~\eqref{eq:prob_kl_rp} is consistent with the traditional policy gradient method shown in eq.~\eqref{eq:prob_policy}, although the sign is inverted due to the difference between maximization and minimization problems.

\subsubsection{Forward KL divergence}

Next, the forward KL divergence is employed for the following minimization problem.
\begin{align}
    \min \mathbb{E}_{p_e}[\mathrm{KL}(\pi^+(a \mid s) \mid \pi(a \mid s)) - \mathrm{KL}(\pi^-(a \mid s) \mid \pi(a \mid s))]
    \label{eq:prob_kl_fp}
\end{align}
Similarly, the gradient of the above two terms inside the expectation over $\phi$, $g_\phi^F$, is derived as follows:
\begin{align}
    g_\phi^F &=
    \int_a
    - \cfrac{p_Q(s, a)}{p_V(s)} b(a \mid s) g_\phi^\pi
    + \cfrac{1 - p_Q(s, a)}{1 - p_V(s)} b(a \mid s) g_\phi^\pi
    da
    \nonumber \\
    &= \mathbb{E}_{b} \left [ - g_\phi^\pi \left \{ \cfrac{p_Q(s, a)}{p_V(s)} - \cfrac{1 - p_Q(s, a)}{1 - p_V(s)} \right \} \right ]
    \nonumber \\
    &= \mathbb{E}_{b} \left [ - g_\phi^\pi \cfrac{p_Q(s, a) - p_V(s)}{p_V(s)(1 - p_V(s))} \right ]
    \nonumber \\
    &\propto \mathbb{E}_{b} \left [ - g_\phi^\pi
    \tau \left ( \cfrac{p_Q(s, a)}{p_V(s)} - 1 \right ) \right ]
    \nonumber \\
    &= \mathbb{E}_{b} \left [ - g_\phi^\pi
    \tau \left \{ \exp \left (\cfrac{Q(s, a) - V(s)}{\tau} \right ) - 1 \right \} \right ]
    \label{eq:grad_kl_fp}
\end{align}
where the proportion is given by multiplying the fourth line by $\tau (1 - p_V(s))$.
Without any assumptions, the same term as in eq.~\eqref{eq:grad_kl_fv} is applied to the gradient (for the value function and the log-likelihood of policy, respectively).
In addition, the expectation for $b(a \mid s)$ is naturally derived without the importance sampling.
In other words, FKL-RL can be regarded to be off-policy learning, which can be learned from data collected independently of $\pi$.

\section{Analysis of differences between RKL-RL and FKL-RL}
\label{sec:analysis}

\subsection{Qualitative differences between forward/reverse KL divergences}

Now, two types of optimization problems using the forward/reverse KL divergences are derived.
We refer to the traditional method shown in eqs.~\eqref{eq:grad_kl_rv_approx} and~\eqref{eq:grad_kl_rp_approx} and the new derivation shown in eqs.~\eqref{eq:grad_kl_fv} and~\eqref{eq:grad_kl_fp} as RKL-RL and FKL-RL to easily distinguish between them.

In general, the difference between forward/reverse KL divergences is explained by the way of fitting for its minimization.
Specifically, when the learning model has little expressive capability (in particular, the lack of multimodality) for the target distribution, minimization of reverse KL divergence makes the learning model fit into a part of the target.
On the other hand, minimization of forward KL divergence tries to fit the learning model on the entire target, resulting in ambiguous model due to the lack of expressive capability.

Due to their properties, given a simple model (e.g. normal distribution), only RKL-RL would acquire one of the local solutions.
This expectation is consistent with the fact that the traditional policy-gradient method only shows convergence to local solutions.
Inspired by this fact, even imitation learning, which minimizes forward KL divergence, has been converted into the problem for minimizing reverse KL divergence~\citep{ke2019imitation,uchibe2020imitation}.
However, it is not clear whether the obtained local solution has sufficient performance.
To find better solutions, the latest RL methods focus on additional regularization problems that make the policy improvement conservative~\citep{schulman2017proximal} or encourage exploration~\citep{haarnoja2018soft}.

On the other hand, FKL-RL with the simple model may not acquire any local solutions.
Instead, it is expected to explore sufficiently to capture the entire target.
The difference in exploration capability affects when the learning model is made more complex like mixture model employed in several practical studies~\citep{kormushev2010robot,daniel2016hierarchical,sasaki2019multimodal}.
That is, RKL-RL may be prone to local solutions, and even if the number of components is increased, there is no guarantee to make the components fit into different solutions.
In contrast, FKL-RL may effectively use all the components to represent the entire target.

Note that the policy optimization in FKL-RL and RKL-RL is not simple minimization problem of forward/reverse KL divergences.
In addition, each update can produce the above properties, but the target distribution changes with each update in RL scheme, inducing the different convergence results.
Therefore, it needs to be investigated whether these behaviors can be expected.

\subsection{Mathematical differences between RKL-RL and FKL-RL}

To clarify the mathematical differences between them, their gradients are summarized as below.
\begin{align}
    &\begin{cases}
        g_{\theta,\phi}^R &= - \mathbb{E}_{p_e, b}[\delta(s, a) (g_\theta^V, \rho(s, a) g_\phi^\pi)]
        \\
        \delta(s, a) &:= Q(s, a) - V(s) \simeq r + \gamma V(s^\prime) - V(s)
        \\
        \rho(s, a) &:= \pi(a \mid s) / b(a \mid s)
    \end{cases}
    \label{eq:grad_rkl} \\
    &\begin{cases}
        g_{\theta,\phi}^F &= - \mathbb{E}_{p_e, b}[\tilde{\delta}(s, a) (g_\theta^V, \tilde{\rho}(s, a) g_\phi^\pi)]
        \\
        \tilde{\delta}(s, a) &:= \tau \left \{ \exp \left (\cfrac{\delta(s, a)}{\tau} \right ) - 1 \right \}
        \\
        \tilde{\rho}(s, a) &:= 1
    \end{cases}
    \label{eq:grad_fkl}
\end{align}
where $(\cdot, \cdot)$ with two gradient vectors denotes the concatenation of them, and $s^\prime$ is sampled from $p_e(s^\prime \mid s, a)$.
Note again that $g_\theta^V = \nabla_\theta V(s)$ and $g_\phi^\pi = \nabla_\phi \ln \pi(a \mid s)$ for simplicity.
$\delta(s, a)$ and $\rho(s, a)$ denote TD error and density ratio in the traditional method.
$\tilde{\delta}(s, a)$ and $\tilde{\rho}(s, a)$ are regarded as the surrogated versions of $\delta(s, a)$ and $\rho(s, a)$ in FKL-RL, respectively.
That is, the differences in RKL-RL and FKL-RL are only in them.
The effects of these two differences on learning are discussed in the following sections.

\subsubsection{Needs of density ratio regularization on RKL-RL}

In RKL-RL (a.k.a. the traditional RL), $\rho(s, a)$ is introduced to change the sampler of action.
Although the importance sampling is theoretically acceptable, in terms of numerical stability, the smaller denominator (i.e. likelihood of the baseline policy $b(a \mid s)$) tends to take the larger $\rho(s, a)$, which would diverge the updates of $\phi$.
It is also known that the approximation performance of expectation with the importance sampling is deteriorated when the alternative sampler (i.e. $b(a \mid s)$) deviates from the original probability (i.e. $\pi(a \mid s)$).

To alleviate these issues in the importance sampling on RKL-RL, the previous studies proposed several methods for regularizing the importance sampling.
For example, \citet{munos2016safe} clipped $\rho(s, a)$ so as to be within $[0, 1]$.
\citet{schulman2017proximal} (and its modified version~\citep{kobayashi2020proximal}) regularized the divergence between $b(a \mid s)$ and $\pi(a \mid s)$ by surrogating $\rho(s, a) g_\phi^\pi$.
Compared to FKL-RL, there is room for improvement in the design of $\rho(s, a)$ as above, but its optimization in terms of learning performance is open yet.

\subsubsection{Optimistic updates on FKL-RL}

\begin{figure}[tb]
    \centering
    \includegraphics[keepaspectratio=true,width=0.95\linewidth]{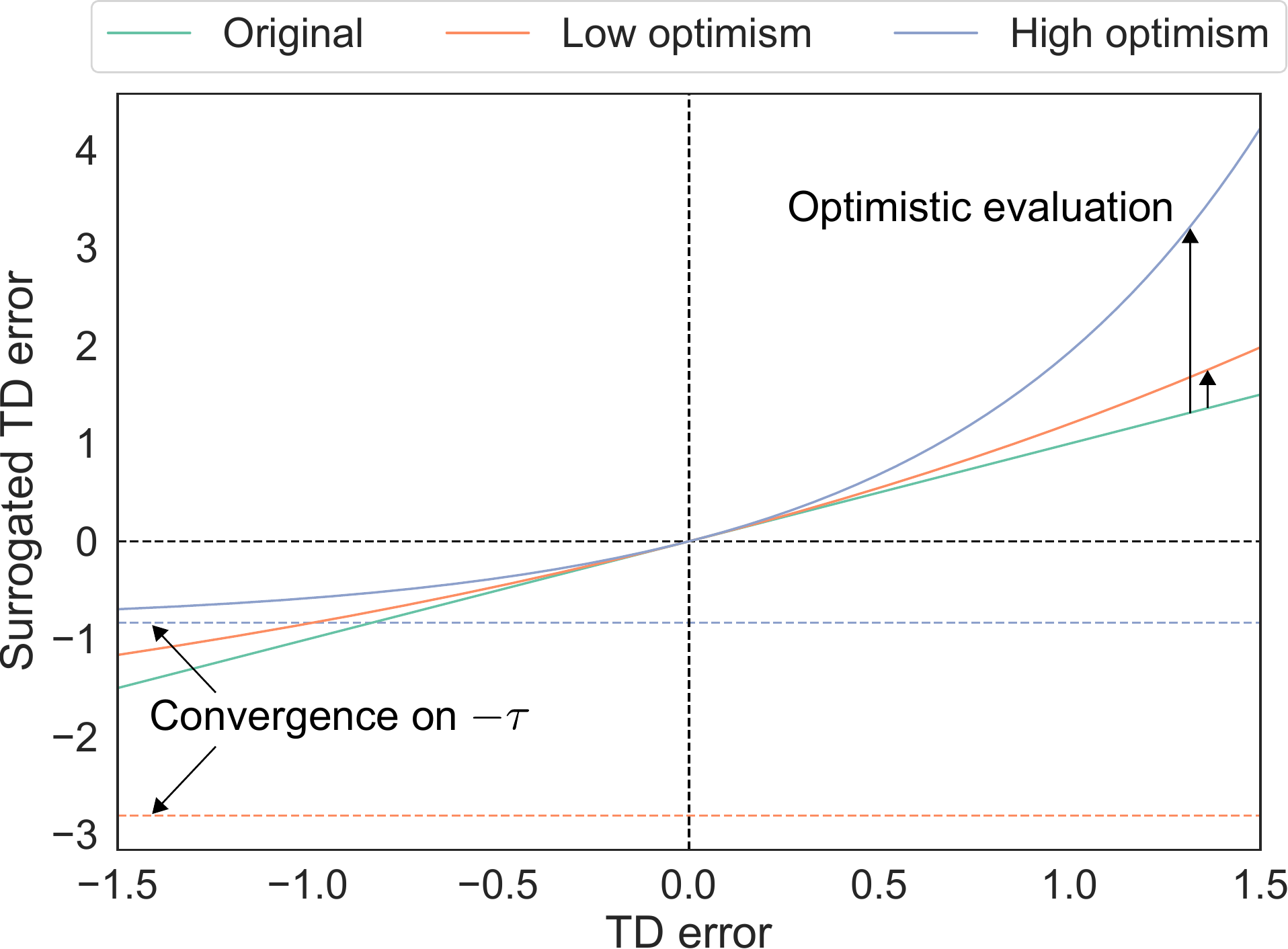}
    \caption{Illustration of (surrogated) TD errors, $\delta$ and $\tilde{\delta}$:
        both $\delta$ and $\tilde{\delta}$ always have the same sign;
        $\tilde{\delta}$ is larger than $-\tau$, two types of which are shown in the dashed lines with the colors corresponding to $\tilde{\delta}$ that have high/low optimism, respectively;
        as $\tau$ becomes smaller, the optimism of $\tilde{\delta}$, i.e. the distance from $\delta$, becomes larger.
    }
    \label{fig:img_tderr}
\end{figure}

$\delta(s, a)$ and $\tilde{\delta}(s, a)$ are illustrated in Fig.~\ref{fig:img_tderr}.
As can be seen in the figure, their signs are reversed at $\delta(s, a) = \tilde{\delta}(s, a) = 0$, and $\tilde{\delta}(s, a)$ is basically larger than $\delta(s, a)$ due to the nonlinearity gained from exponential function.

This nonlinearity leads to optimistic updates on FKL-RL.
Mathematically, $\tilde{\delta}(s, a)$ has analytical lower bound: $\inf \tilde{\delta}(s, a) = - \tau$.
In addition, although $\mathbb{E}_{\pi}[\delta(s, a)] = 0$ since $\mathbb{E}_{\pi}[Q(s, a)] = V(s)$, $\mathbb{E}_{\pi}[\tilde{\delta}(s, a)]$ is non-negative as below.
\begin{align}
    \mathbb{E}_{\pi}[\tilde{\delta}(s, a)] &= \tau \mathbb{E}_{\pi}\left [\exp \left (\cfrac{\delta(s, a)}{\tau} \right ) \right ] - \tau
    \nonumber \\
    &\geq \tau \exp \mathbb{E}_{\pi}\left [\cfrac{\delta(s, a)}{\tau} \right ] - \tau = 0
\end{align}
where Jensen's inequality is applied to swap exponential and expectation operations.
Such positively biased $\tilde{\delta}(s, a)$ would make the value function and the policy optimistically update by acting on their gradients.
Note that although the expectation is indeed non-negative, there is clearly only one equilibrium point which is $Q(s, a) = V(s)$, hence $V(s)$ is expected to converge to $Q(s, a)$ after asymmetric updating.
In addition, the update of $\pi$ will also stop when $Q(s, a) = V(s)$, resulting in the convergence capability to one of the local solutions with $Q(s, a) = V(s)$.

Optimism in RL has long been discussed as a concept to resolve the exploration-exploitation dilemma, and implemented like by optimistic initialization~\citep{machado2015domain,rashid2019optimistic}; and increasing the value of exploration using upper confidence~\citep{sunehag2015rationality,curi2020efficient}.
Recently, optimistic decision making in organisms has been modeled by optimistic RL with asymmetric learning rates and validated~\citep{lefebvre2017behavioural}.
In other words, the optimistic updates on FKL-RL have the potential to escape from local solutions and to find global solution efficiently.
That is consistent with the behavior of forward KL divergence minimization, which tries to capture the entire target distribution.

\subsection{Adaptive design of uncertainty parameter as optimism parameter}

One of the main differences in RKL-RL and FKL-RL is the uncertainty parameter, $\tau$, which is left only in FKL-RL.
As can be seen in Fig.~\ref{fig:img_tderr}, $\tau$ can be interpreted as a coefficient that determines optimism, although the quantitative relationship between them is hard to understand.
To intuitively and adaptively design $\tau$, a heuristic conversion of $\tau$ is introduced.

The important fact is that $\tilde{\delta}$ has the lower bound $\inf \tilde{\delta}(s, a) = - \tau$, and $\tilde{\delta}$ would not change much in the vicinity of this bound, making it difficult to grasp differences in solutions.
If we know the scale of $\delta$ as $\Delta$, we can design $\tau$ by specifying how close $\tilde{\delta}(-\Delta)$ is to the lower bound.
Given an alternative parameter $\eta \in (0, 1)$, the following design of $\tau$ is derived.
\begin{align}
    \tau \left \{ \exp \left (\cfrac{-\Delta}{\tau} \right ) - 1 \right \} &= - \eta \tau
    \nonumber \\
    \exp \left (\cfrac{-\Delta}{\tau} \right ) &= 1 - \eta
    \nonumber \\
    \tau &= - \cfrac{\Delta}{\ln (1 - \eta)}
    \label{eq:tau_adapt}
\end{align}
In this design, $\eta$ is given as a parameter that is more intuitively related to optimism:
if $\eta \simeq 0$, the optimistic effect from the exponential function will be vanished, and vice versa.
In addition, $\Delta$ is expected to be task- and algorithm-dependent, but with this design, there is no need to tune $\tau$ for each task with each algorithm.

Instead, $\Delta$ for the specific task solved by the specific algorithm is unknown before solving it.
In other words, $\Delta$ needs to be estimated during training.
The estimation is heuristically given as follows:
\begin{align}
    \Delta_\mathrm{max} &\gets \max(\beta \Delta_\mathrm{max}, \max(|\delta|))
    \label{eq:err_max} \\
    \Delta &\gets \beta \Delta + (1 - \beta) \Delta_\mathrm{max}
    \label{eq:err_scale} \\
    \tau &= - \cfrac{\max(\min(\Delta, \epsilon^{-1}), \epsilon)}{\ln (1 - \eta)}
    \label{eq:tau_adapt2}
\end{align}
where $\Delta_\mathrm{max}$ stores the recent maximum magnitude of $\delta$, then it is smoothly transferred into $\Delta$.
Note that the initial value of $\Delta_\mathrm{max}$ is given to be $\epsilon^{-1}$ as the highest uncertainty, resulting in almost no optimism at the beginning of learning.
$\beta \simeq 1$ is for slowly decaying the past large $\delta$ and updating $\Delta$ toward $\Delta_\mathrm{max}$.
Since max operator switches the output value discretely, this two-step update provides smooth estimation of $\Delta$ and suppresses update fluctuations.
In addition, for numerical stability, $\Delta$ is clamped using $\epsilon \ll 1$.

\section{Integration with methods to accelerate learning}
\label{sec:integration}

\subsection{Prioritized experience replay}

The recent standard to accelerate learning is to reuse previous experiences, so-called experience replay~\citep{lin1992self}.
Specifically, it stores Markovian data, $(s, a, s^\prime, r)$, to a replay buffer with $N_c$ capacity.
After finishing each episode, it samples $N_b$ data from the buffer randomly to compute the gradients of parameters and to update them, repeating this process $N_e$ times.
As a sampling method, prioritized experience replay (PER)~\citep{schaul2015prioritized}, which prioritizes the reuse of important data, has been proposed.
PER defines the sampling probability of $i$-th tuple, $p_i$, based on the magnitude of TD error.
\begin{align}
    w_i = |\delta_i| + \epsilon
    , \
    p_i = \cfrac{w_i^{\alpha_P}}{\sum_i w_i^{\alpha_P}}
    \label{eq:per_weight_origin}
\end{align}
where $\epsilon$ denotes the small value to ensure that $p_i > 0$ and $\alpha_P \geq 0$ is for adjusting the influence of weights.
For simplicity, the arguments for TD error, $(s, a)$ are omitted and identified by a subscript $i$.
Note that this weighted sampling can be regarded as importance sampling against uniform distribution, hence in the original paper, the density ratio surrogated by the hyperparameter $\beta_P \in [0, 1]$ is multiplied by the loss (or gradient) corresponding to each sample.

In the traditional RL (and RKL-RL), TD error is symmetric, so that both positive and negative values are considered equally important.
In FKL-RL, however, we have a new option: instead of $\delta$ in eq.~\eqref{eq:grad_rkl}, $\tilde{\delta}$ in eq.~\eqref{eq:grad_fkl} can be used for importance.
\begin{align}
    \tilde{w}_i &= |\tilde{\delta}_i| + \epsilon
    , \
    p_i = \cfrac{\tilde{w}_i^{\alpha_P}}{\sum_i \tilde{w}_i^{\alpha_P}}
    \label{eq:per_weight_fkl}
\end{align}
This simple modification indicates that better results are actively reused due to the asymmetry of $\tilde{\delta}$.
As a result, a function similar to self-imitation learning (SIL)~\citep{oh2018self}, which facilitates exploration by excluding updates with negative TD errors during replay, can be expected.

\subsection{Eligibility traces}

When FKL-RL is integrated with $\lambda$-return and eligibility traces~\citep{sutton2018reinforcement}, we have to consider the fact that the value function in FKL-RL is contained in nonlinear (i.e. exponential) function.
To clarify this problem, the definition of generalized advantage estimation (GAE) by~\citet{schulman2015high}, which is equivalent with $\lambda$-return, is first introduced.
\begin{align}
    \delta_t^\lambda := \sum_{k=0}^\infty (1 - \lambda) \lambda^k \delta_{t+k}
    \label{eq:def_gae}
\end{align}
where $\lambda \in [0, 1]$ denotes the hyperparameter for balancing tradeoff between bias and variance.
Originally, $\delta^\lambda$ replaces $\delta$ in eq.~\eqref{eq:grad_rkl}, and in that time, its backward view can be obtained theoretically since $\delta^\lambda$ is a linear weighted summation.
With this feasibility of the backward view, the eligibility traces method is derived:
it accumulates the gradients (before multiplying by $\delta$) at each time as traces while decaying them, and defines new gradients for update by multiplying the traces and $\delta$ at the current time.

However, if $\delta^\lambda$ alternates $\delta$ in eq.~\eqref{eq:grad_fkl}, the summation operation is contained in the exponential function.
In that case, GAE itself is naively computable, but its backward view cannot be obtained due to the nonlinearity of the exponential function.
Therefore, the eligibility traces method cannot be applied to FKL-RL in the current form.

To resolve this problem, we can focus on the fact that GAE can be interpreted as the expectation of $\delta_{t+k}$ when $k$ is sampled from geometric distribution with parameter $1 - \lambda$.
That is, Jensen's inequality can be applied as follows:
\begin{align}
    &\delta_t^\lambda = \tau \cfrac{\delta_t^\lambda}{\tau}
    = \tau \ln \exp \cfrac{\delta_t^\lambda}{\tau}
    \nonumber \\
    &\leq \tau \ln \sum_{k=0}^\infty (1 - \lambda) \lambda^k
    \exp \left ( \cfrac{\delta_{t+k}}{\tau} \right )
    \label{eq:ineq_gae}
\end{align}
Using this inequality, substituting $\delta_t^\lambda$ for $\delta$ in eq.~\eqref{eq:grad_fkl} yields the following relationship.
\begin{align}
    &\tau \left \{ \exp \left (\cfrac{\delta_t^\lambda}{\tau} \right ) - 1 \right \}
    \nonumber \\
    &\leq \tau \left [ \exp \left \{ \cfrac{1}{\tau}
    \tau \ln \sum_{k=0}^\infty (1 - \lambda) \lambda^k
    \exp \left ( \cfrac{\delta_{t+k}}{\tau} \right ) \right \} - 1 \right ]
    \nonumber \\
    &= \tau \left \{ \sum_{k=0}^\infty (1 - \lambda) \lambda^k
    \exp \left ( \cfrac{\delta_{t+k}}{\tau} \right ) - 1 \right \}
    \nonumber \\
    &= \sum_{k=0}^\infty (1 - \lambda) \lambda^k
    \tau \left \{ \exp \left ( \cfrac{\delta_{t+k}}{\tau} \right ) - 1 \right \}
    \nonumber \\
    &= \sum_{k=0}^\infty (1 - \lambda) \lambda^k \tilde{\delta}_{t+k}
    =: \tilde{\delta}_t^\lambda
    \label{eq:ineq_et}
\end{align}
As can be seen from the last equation, the surrogated GAE with $\tilde{\delta}_{t+k}$ is derived as the upper bound of $\tilde{\delta}$ with the original GAE.
Since $\tilde{\delta}^\lambda$ is defined as a linear weighted summation as well as $\delta^\lambda$, its backward view can be obtained.
Therefore, with $\tilde{\delta}^\lambda$, the eligibility traces method can be applied even to FKL-RL.
As a remark, when $\tilde{\delta}_t^\lambda$ is utilized for updates, optimistic learning occurs here as well since it is the upper bound.

\section{Simulation}
\label{sec:simulation}

\subsection{Conditions}

\begin{figure}[tb]
    \centering
    \includegraphics[keepaspectratio=true,width=0.95\linewidth]{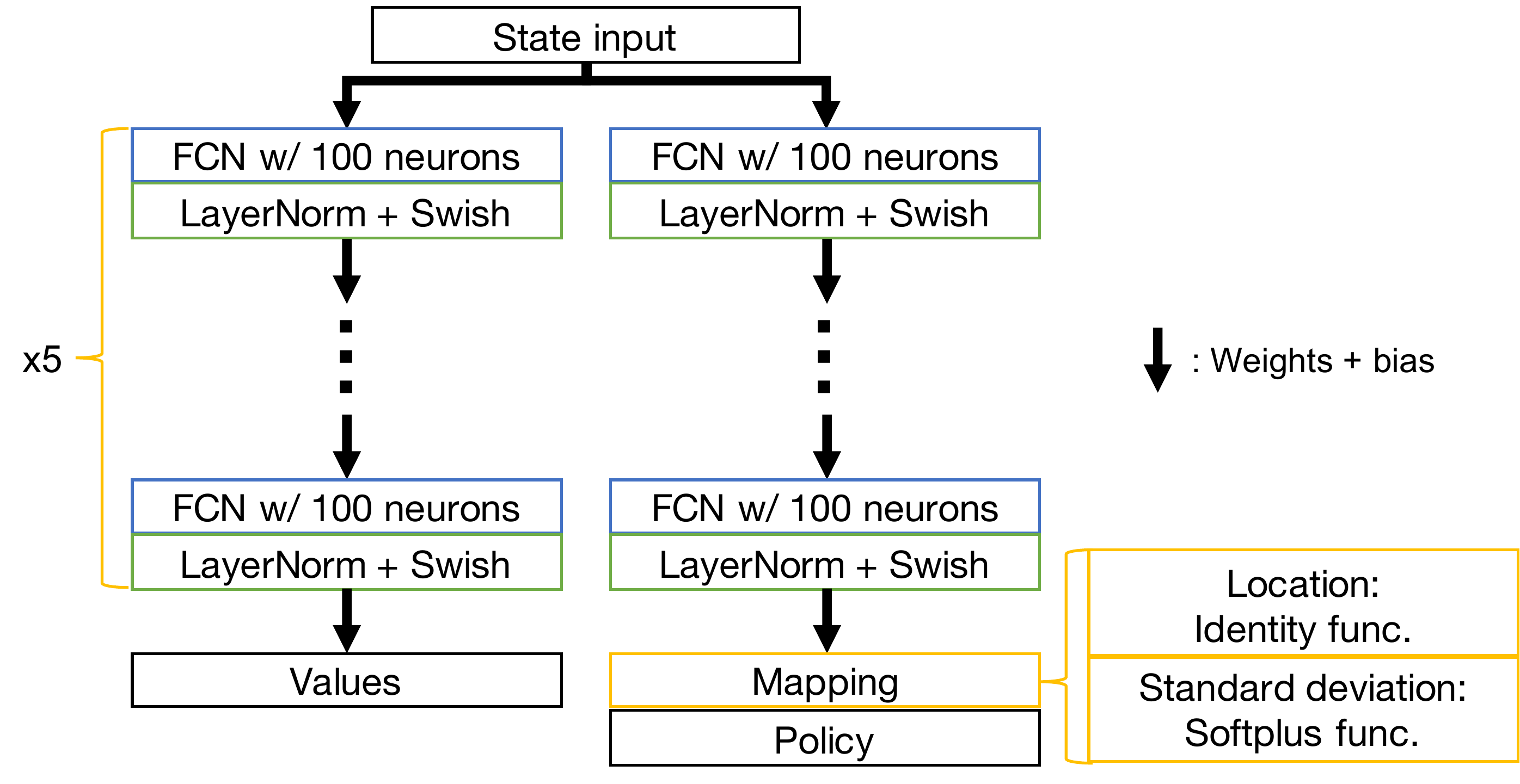}
    \caption{Network architectures for approximating the value function $V$ and the policy $\pi$:
        both functions basically use the same architecture with units of a fully connected layer, layer normalization, and swish activation function;
        the outputs from the five units are transformed into $V$ and $\pi$ via a linear transformation by weights and bias and the respective mapping functions.
    }
    \label{fig:network_architecture}
\end{figure}

\begin{table*}[tb]
    \caption{Common parameters: the two values in curly brackets are those used in the first and second experiments, respectively.}
    \label{tab:parameter}
    \centering
    \begin{tabular}{ccc}
        \hline\hline
        Symbol & Meaning & Value
        \\ \hline
        $\gamma$ & Discount factor & 0.99
        \\
        $\alpha$ & Learning rate & \{$5 \times 10^{-4}$, $5 \times 10^{-6}$\}
        \\
        $\epsilon$ & Minimal value for computational stabilization & $10^{-5}$
        \\
        $\beta$ & Decaying factor for estimation of $\Delta$ & 0.999
        \\
        $\tau_H$ & For entropy maximization~\citep{shi2019soft} & 0.1
        \\
        $(\tau_S, \nu_S)$ & For t-soft update~\citep{kobayashi2021t} & (0.5, 1.0)
        \\
        $(\lambda_\mathrm{max}^1, \lambda_\mathrm{max}^2, \kappa)$ & For alternative eligibility traces~\citep{kobayashi2020adaptive} & (0.5, 0.9, 1.0)
        \\
        $(N_c, N_e, N_b, \alpha_P, \beta_P)$ & For prioritized experience replay~\citep{schaul2015prioritized} & ($10^5$, 32, 32, 0.6, 0.4)
        \\
        \hline\hline
    \end{tabular}
\end{table*}

The performances of FKL-RL and RKL-RL are investigated through numerical simulations by Pybullet on OpenAI Gym framework~\citep{brockman2016openai,coumans2016pybullet}.
At a first experiment, \textit{InvertedDoublePendulumBulletEnv-v0} (abbreviated as DoublePendulum) and \textit{HopperBulletEnv-v0} (abbreviated as Hopper) are tried.
The higher return may be gained from them as the higher optimism is set since their global solutions exist beyond the off-balance behaviors.
In addition, for more realistic robot simulation, \textit{MinitaurBulletEnv-v0} (abbreviated as Minitaur) is also tried as a second experiment.
Note that its default configurations are modified to be more realistic (see \ref{app:minitaur}).

The network architecture for approximating the value function and the policy is designed using PyTorch~\citep{paszke2017automatic}, as illustrated in Fig.~\ref{fig:network_architecture}.
Although the two functions are approximated by separate networks, they share the same basic architecture except for their outputs.
Since all the inputs are real values, five fully connected networks (FCNs) are given as hidden layers.
Each hidden layer has 100 neurons, and maps to nonlinear space through Layer Normalization and Swish activation function~\citep{ba2016layer,elfwing2018sigmoid}.
Finally, the networks for the value function output five bootstrapped real values, each of which is used for computing $\delta$ and averaged.
The networks for the policy output a diagonal normal distribution.
These networks are optimized by t-Yogi, which integrates Yogi~\citep{zaheer2018adaptive} with t-momentum~\citep{ilboudo2020robust}, with default configurations except learning rate.

The entropy term of the policy is added to reward as a bonus to stabilize learning, which is inspired by SAC~\citep{haarnoja2018soft} and soft policy gradient~\citep{shi2019soft}: more specifically, $r \gets r - \tau_H \ln \pi(a \mid s)$ with $\tau_H \geq 0$ gain.
Note that this implementation is heuristic and different from entropy-regularized RL in FKL-RL, that is, the theoretically-rigorous implementation of the entropy-regularized FKL-RL is open yet.
In addition, the target network is introduced for all the networks and updated by t-soft update~\citep{kobayashi2021t}.
The baseline policy $b(a \mid s)$ is therefore set as the output from the target network.
Since it has been reported that the standard eligibility traces method would not perform well with nonlinear function approximation~\citep{van2016effective,kobayashi2020adaptive}, its variant~\citep{kobayashi2020adaptive}, which alleviates this problem by adaptively accumulating gradients, is employed.
Parameters used in these implementations are listed up in Table~\ref{tab:parameter}.
Here, the default values are mostly utilized from the previous studies, except the buffer size and the number of drawing from the buffer are reduced due to the limitation of computational resources.
Note that the expectation operations introduced above are approximated by Monte Carlo method (i.e. means of data sampled from probabilities listed in subscripts).
The implementation of the above algorithm is described in \ref{app:algorithm}.

\subsection{Results}

\begin{figure*}[tb]
    \centering
    \subfigure[Normalized score]{
        \includegraphics[keepaspectratio=true,width=0.31\linewidth]{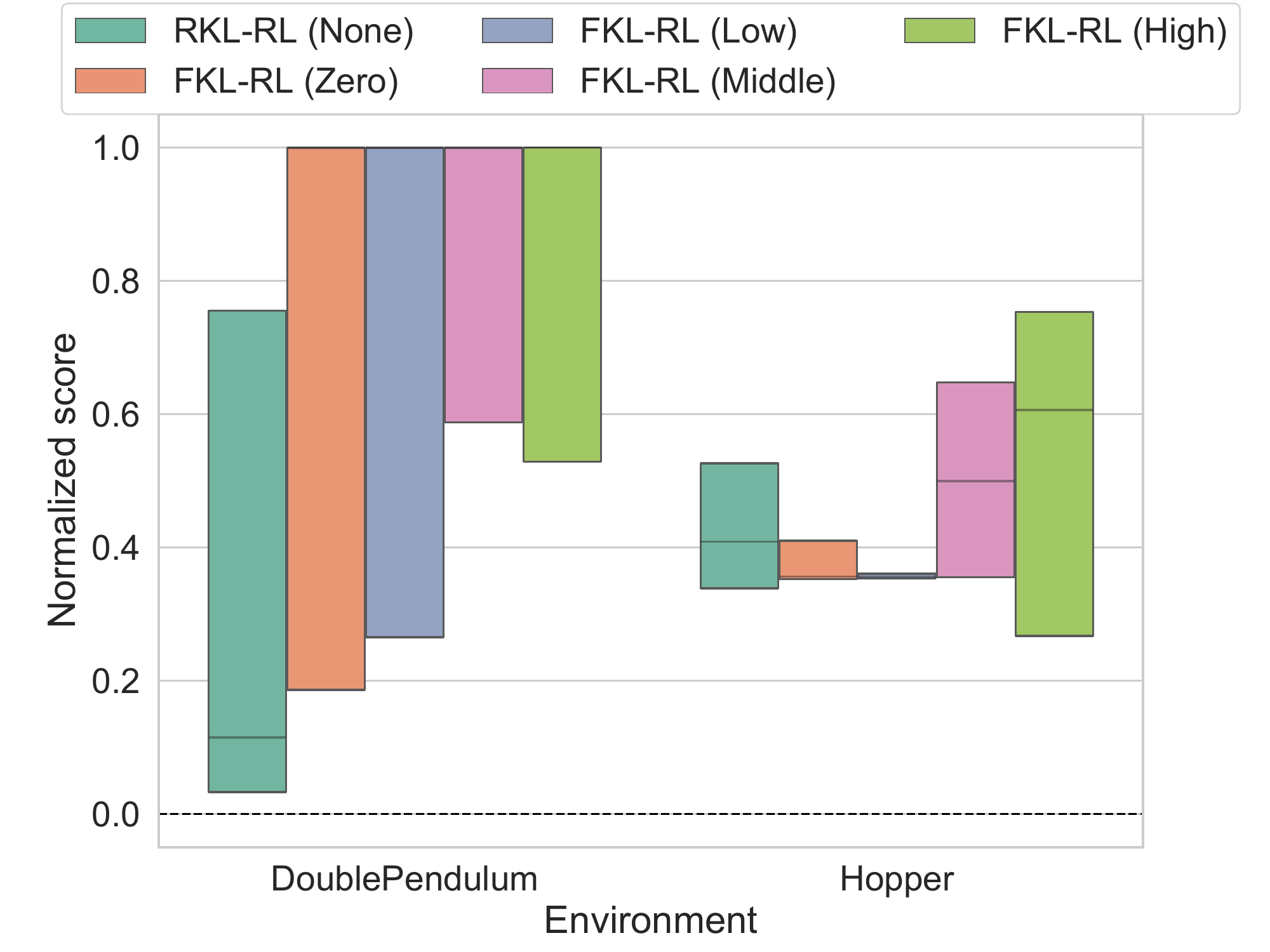}
    }
    \centering
    \subfigure[Absolute TD error]{
        \includegraphics[keepaspectratio=true,width=0.31\linewidth]{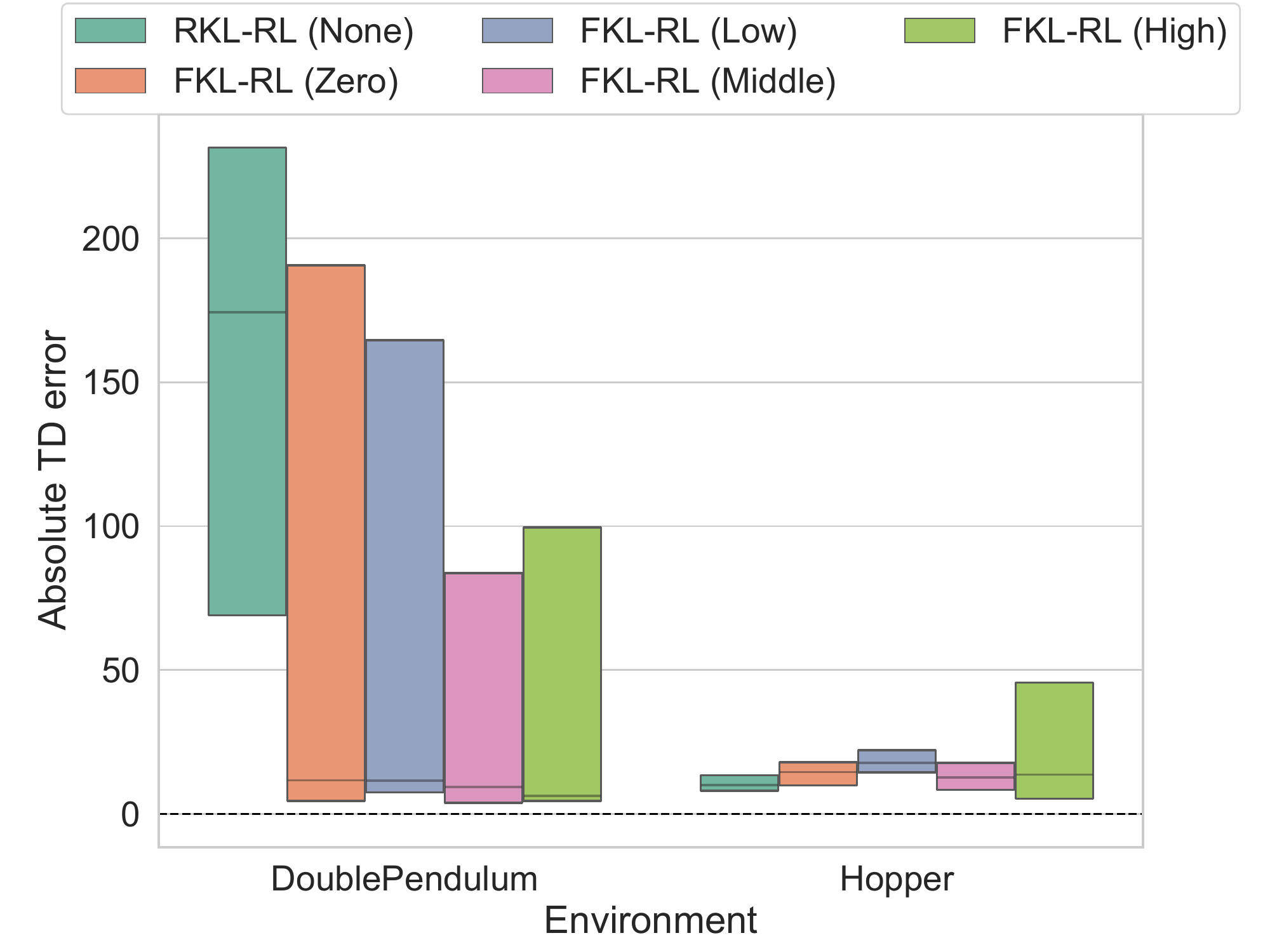}
    }
    \centering
    \subfigure[Log-likelihood of policy]{
        \includegraphics[keepaspectratio=true,width=0.31\linewidth]{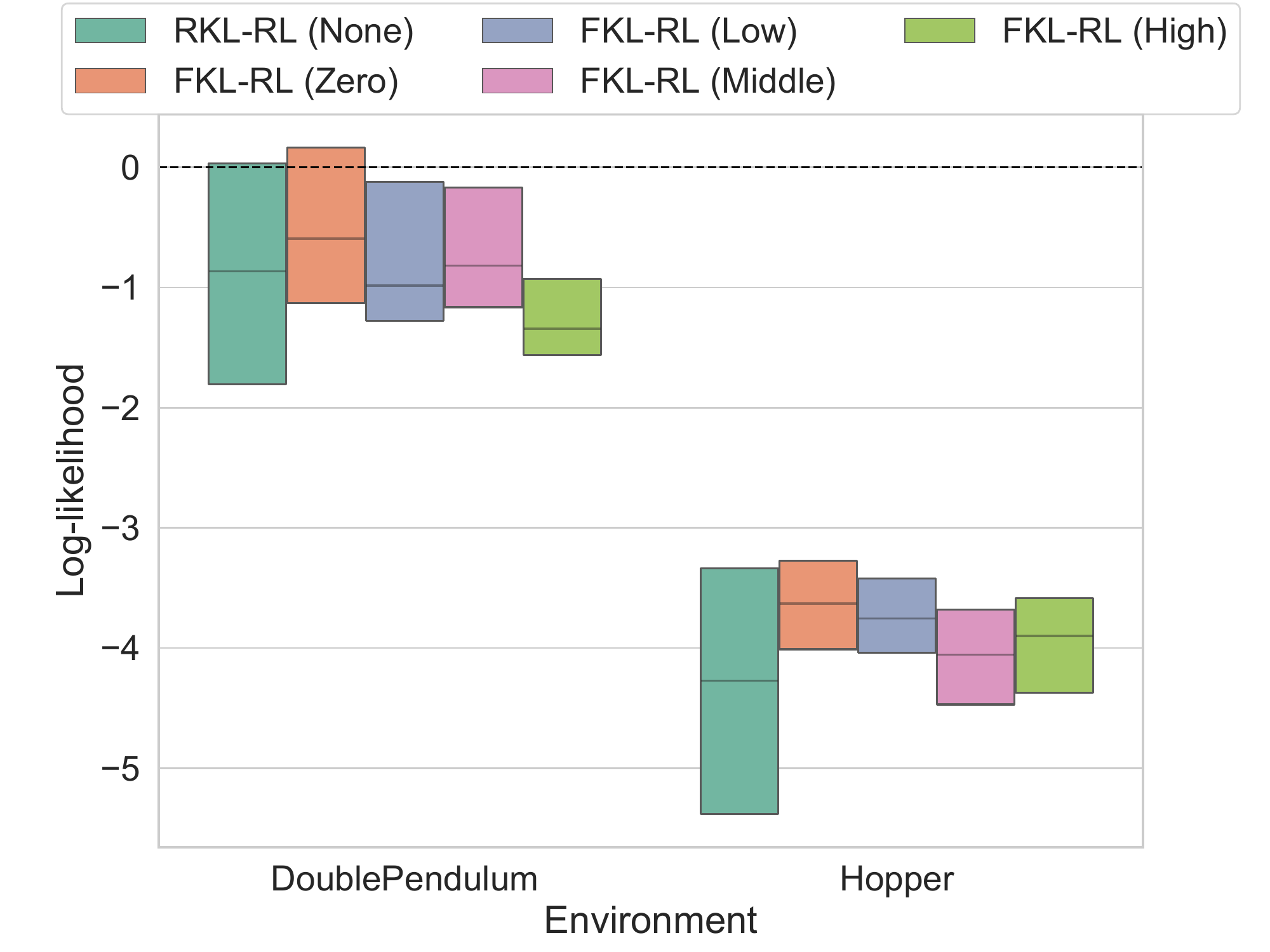}
    }
    \caption{Learning results for the first experiment as boxen plots:
        all data indicated that RKL-RL failed to learn the given tasks due to numerical instability of importance sampling;
        (a) normalized sum of rewards can be increased as the optimism $\eta$ is increased;
        (b) $|\delta|$ can be reduced correctly even in FKL-RL learned by $\tilde{\delta}$, although too much optimism deteriorated the learning accuracy of the value function;
        (c) the higher optimism yielded the smaller $\ln \pi$, and therefore, the stochastic exploration capability was remained until the end.
    }
    \label{fig:sim1_test}
\end{figure*}

\begin{figure}[tb]
    \centering
    \includegraphics[keepaspectratio=true,width=0.95\linewidth]{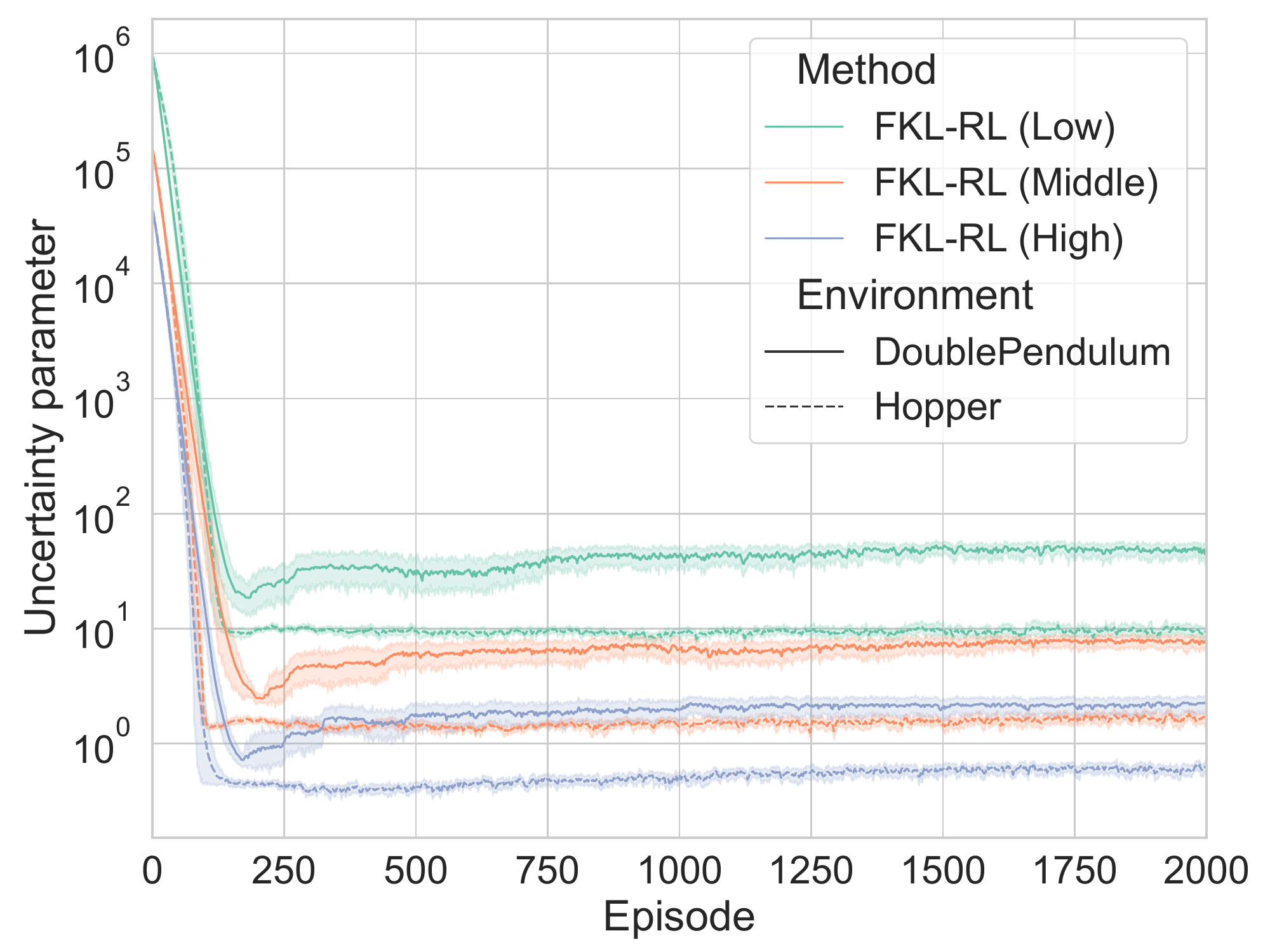}
    \caption{Trajectories of $\tau$ in FKL-RL:
        according to the given optimism $\eta$ and the solved tasks, the respective methods obtained the different stationary $\tau$;
        although the destination of convergence is different, all of them were gained around 250 episodes.
    }
    \label{fig:sim1_temp}
\end{figure}

\subsubsection{First experiment}

As comparisons in the first experiment, the following five configurations are evaluated.
\begin{itemize}
    \item RKL-RL (None) as the baseline
    \item FKL-RL (Zero) with $\tau \to \infty$ for no optimism
    \item FKL-RL (Low) with $\eta = 0.1$ for low optimism
    \item FKL-RL (Middle) with $\eta = 0.5$ for middle optimism
    \item FKL-RL (High) with $\eta = 0.9$ for high optimism
\end{itemize}
By comparing RKL-RL (None) and FKL-RL (Zero), we can see the impact of not having to use importance sampling.
By comparing RKL-RL (None) and FKL-RL (High), both of which have the small $\tau$, the effect of asymmetry can be checked on the approximately aligned definition of optimality.
Note that if $\tau \to 0$ is set in FKL-RL, $\tilde{\delta}$ diverges when $\delta > 0$, which is infeasible in numerical calculations, and therefore, it is excluded.
By comparing four FKL-RLs, the best balance of the optimism can be found.

Both of DoublePendulum and Hopper were solved by 2000 episodes.
In order to evaluate each method with 20 different random seeds, each learned agent is tested 100 times against different initial states.
The normalized score (i.e. the sum of rewards), the absolute TD error $|\delta|$, and the log-likelihood of the policy $\ln \pi$ in that test are summarized in Fig.~\ref{fig:sim1_test}.
In addition, the convergence of $\tau$ in the respective methods and environments is shown in Fig.~\ref{fig:sim1_temp}.

First, we can confirm from Fig.~\ref{fig:sim1_temp} that all of cases are sufficiently stationary after roughly 250 episodes, although they converged to different values depending on the specified optimism $\eta$ and environment (with the specific reward scale).
That is, the proposed method can adaptively adjust $\tau$ to obtain the specified optimism even when the environment to be solved is changed.

Returning to the learning results in Fig.~\ref{fig:sim1_test}, RKL-RL (None) mostly failed to solve the tasks, and its results were strongly different from the others.
By comparing FKL-RL (Zero), we can say that this difference is caused by the existence of importance sampling, which causes numerical instability as mentioned before.

In other results, the learning performance was lower when the optimism is small.
This may be due to the lack of optimistic exploration, which leads to a local solution.
In contrast, FKL-RL (High) with the maximum optimism obtained smaller $\ln \pi$ than them, suggesting that the stochastic exploration capability was remained until the end of learning.
As a result, FKL-RL (High) achieved the higher scores in all the methods.
However, too much optimism deteriorated the learning accuracy of the value function, and according to its inaccurate guide for the policy updates, the variance of scores increased compared to FKL-RL (Middle).
Note that FKL-RL obtained at least local solutions, which implies that it is different from the general fitting by FKL divergence, namely it only inherits the property to optimistically promote the exploration in the entire space.

This result may be related to the fact that the policy-gradient method without the baseline term has a large learning variance~\citep{greensmith2004variance}.
That is, as optimism increases, the lower bound of $\tilde{\delta}$, $-\tau$, becomes closer to zero.
As a result, most of the information in the direction of worsening the policy, which is not contained in the case without the baseline term~\citep{greensmith2004variance}, would be lost.

Overall, FKL-RL (Middle) with a moderate optimism achieved the stable performance improvement from RKL-RL and FKL-RL (Zero) without optimism, although the maximum score on Hopper was inferior to FKL-RL (High).

\subsubsection{Second experiment}

\begin{figure*}[tb]
    \centering
    \subfigure[Score]{
        \includegraphics[keepaspectratio=true,width=0.31\linewidth]{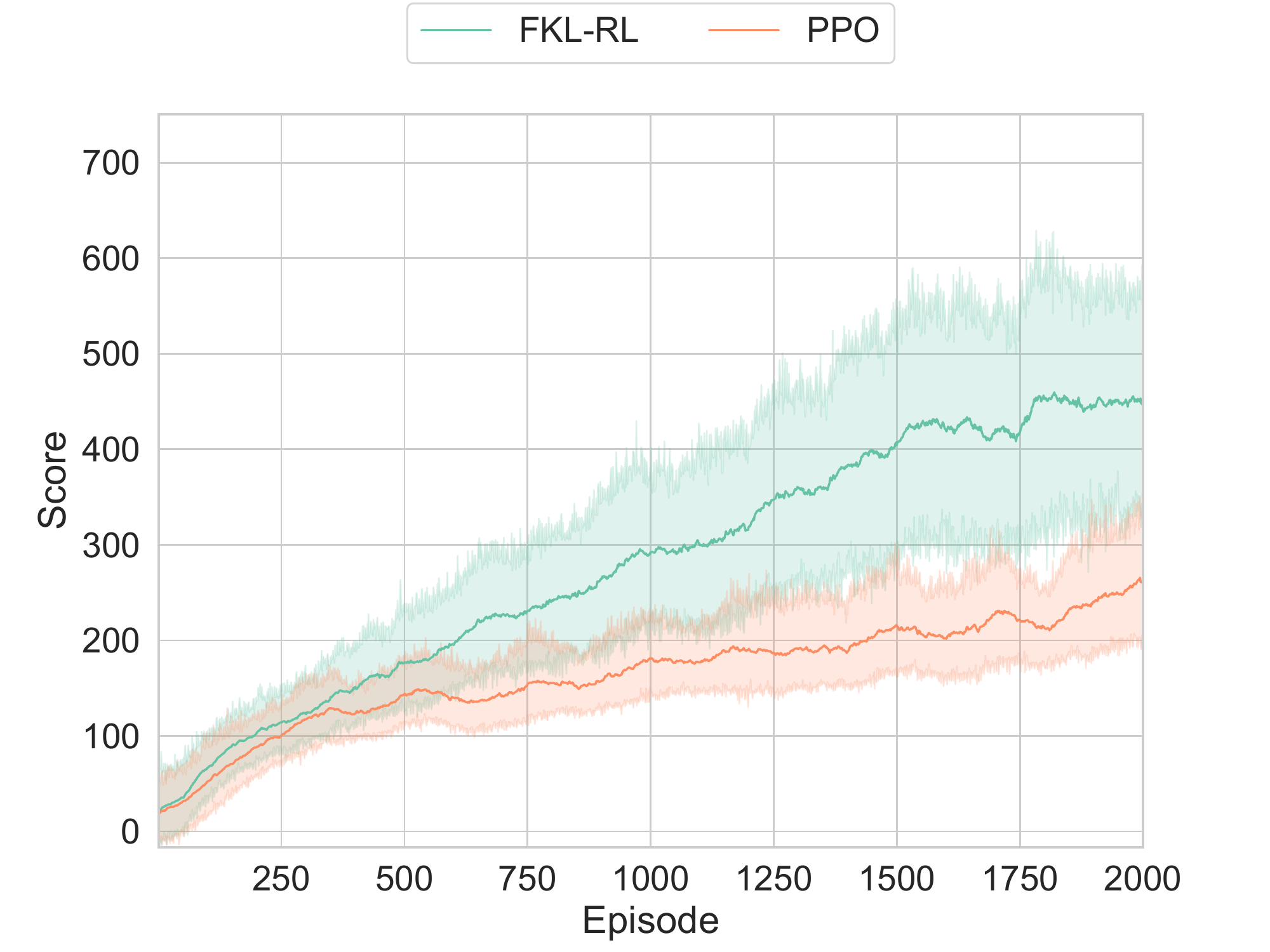}
    }
    \centering
    \subfigure[Absolute TD error]{
        \includegraphics[keepaspectratio=true,width=0.31\linewidth]{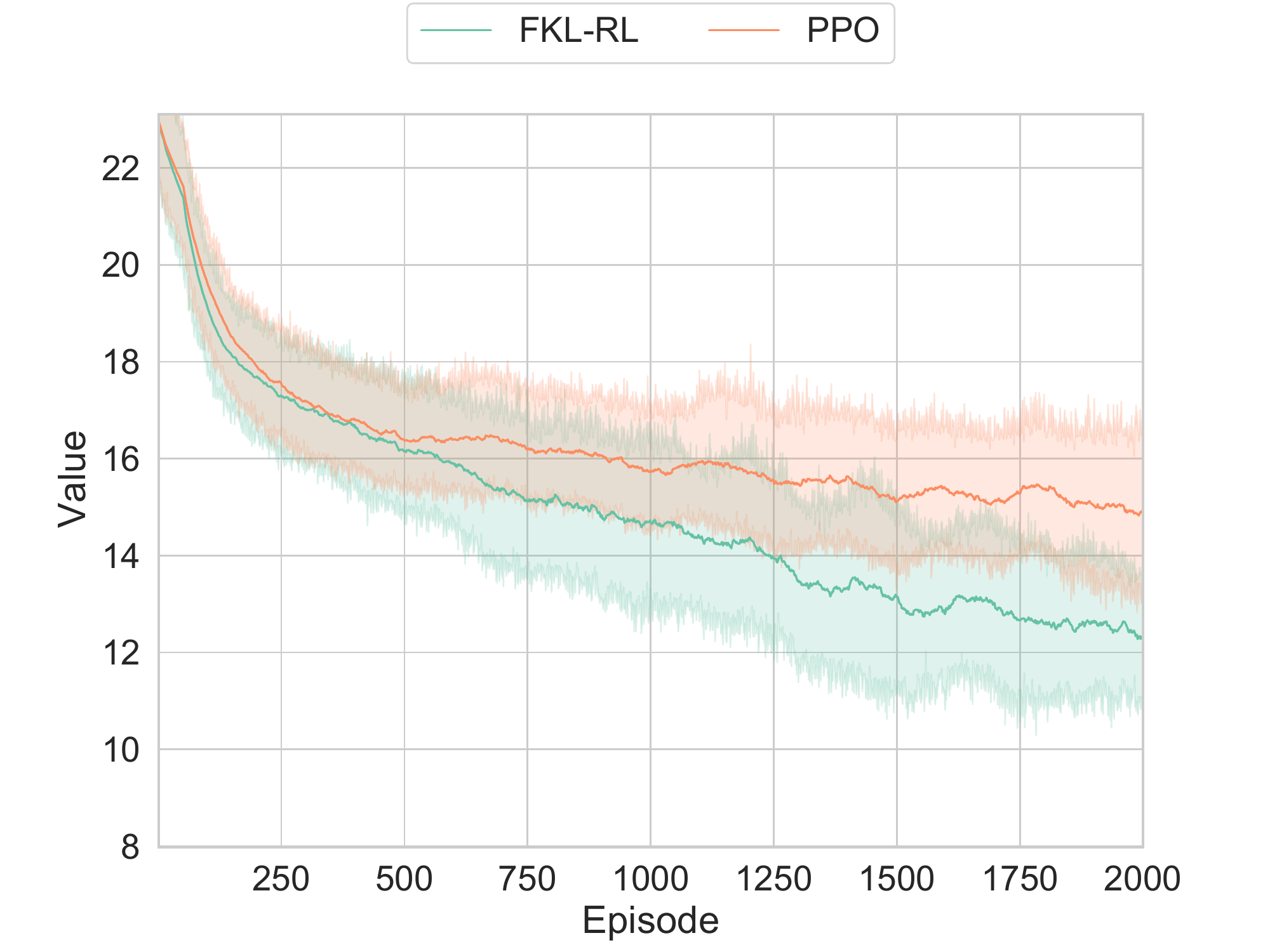}
    }
    \centering
    \subfigure[Log-likelihood of policy]{
        \includegraphics[keepaspectratio=true,width=0.31\linewidth]{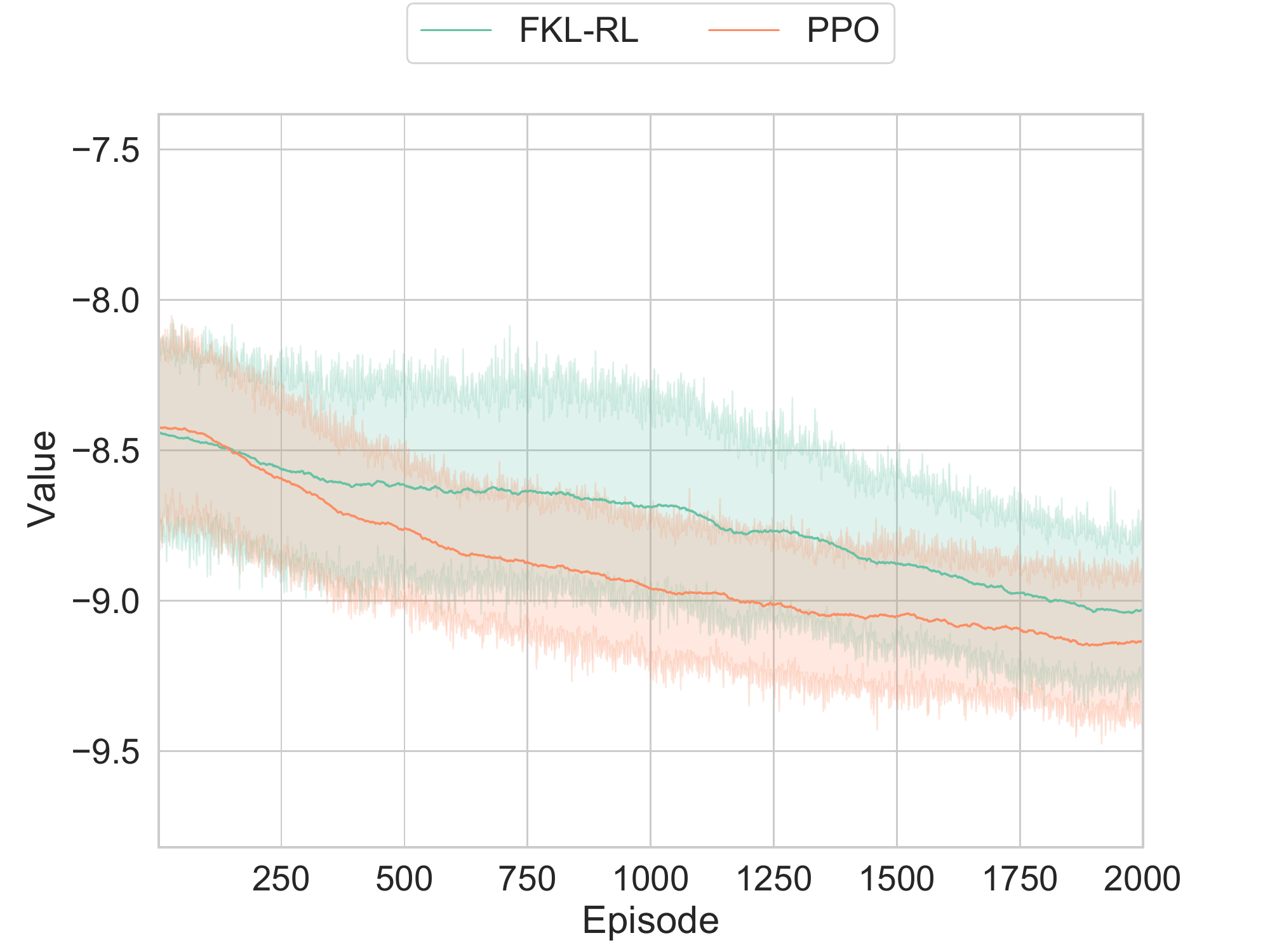}
    }
    \caption{Learning curves for the second experiment:
        shaded area denotes 95~\% confidence interval;
        in (a) the sum of rewards, FKL-RL outperformed PPO in this Minitaur task;
        (b) $|\delta|$ was reduced more efficiently by FKL-RL learned with $\tilde{\delta}$ probably due to its moderate optimism;
        (c) shows that PPO prioritized exploration from the early stage of learning, although that made learning too conservative and caused the failure of increasing the score.
    }
    \label{fig:sim2_learn}
\end{figure*}

As comparisons in the second experiment, the following two configurations are evaluated.
\begin{itemize}
    \item FKL-RL with $\eta = 0.5$ for the best optimism
    \item PPO~\citep{schulman2017proximal} as the baseline
\end{itemize}
By this comparison, FKL-RL will be verified to be the one comparable to the state-of-the-art RL method, which is the same implementation used in \citet{kobayashi2020proximal} with the same parameters as FKL-RL.

Minitaur was solved by 2000 episodes with 10 different random seeds.
Note that, due to the complexity of Minitaur, the learning rate was reduced from $5 \times 10^{-4}$ to $5 \times 10^{-6}$.
Learning curves of the sum of rewards as the score, the absolute TD error $|\delta|$, and the log-likelihood of the policy $\ln \pi$ are depicted in Figs.~\ref{fig:sim2_learn}(a)--(c), respectively.
In addition, almost the best episodes for PPO and FKL-RL are found in an attached video, although their behaviors are not smooth due to observation noise, imperfect PD control, and stochastic behavior.

As can be seen in Fig.~\ref{fig:sim2_learn}(a), FKL-RL succeeded in acquiring the forward movement to some extent, while PPO failed.
The failure of PPO can be attributed to PER with the smaller buffer size and number of uses ($N_c=10^{5}$ and $N_e=32$) than typical use cases ($N_c \simeq 10^{6}$ and $N_e > 100$).
As mentioned above, FKL-RL makes PER more optimistic, which means that the opportunity to learn with less data was effectively utilized in FKL-RL.

Fig.~\ref{fig:sim2_learn}(b) shows that FKL-RL has smaller TD error, in other words, it learned the value function more accurately.
In addition, the randomness of policy was greater in PPO at the early stage of learning, suggesting that PPO improved the policy more conservatively, while FKL-RL gave relatively high priority to exploitation.
These results suggest that FKL-RL was able to estimate the states that would be more valuable earlier than PPO, and prioritize transitions to them, resulting in higher learning efficiency.
Note that, in 7 out of 10 random seeds, the probability of the score exceeding 1000 during learning was high, implying that FKL-RL is not prone to getting stuck in local solutions by prioritizing immediate profits.

\subsection{Discussion}\label{sec:discussion}

When deriving the optimization problem for policy learning, this paper is inspired by the concept of triplet loss (see again the section~\ref{subsec:opt_policy}).
This gives us a learning rule that is approximately equivalent to the traditional RL when using reverse KL divergence, but open issues with triplet loss are left.
One of them is that the policy to be optimized, $\pi$, is not necessarily placed between optimal and non-optimal policies, $\pi^+$ and $\pi^-$.
If the gradients away from $\pi^-$ is stronger than the gradients towards $\pi^+$, $\pi$ will fail to be optimized.
This may be one of the factors that make the policy gradient method unstable.
In fact, in the above simulations, the success or failure of learning was affected by random seeds, and this instability has been pointed out in previous studies~\citep{colas2018many,henderson2018deep}.

Only if $\pi$ is sufficiently close to the baseline policy $b$, this problem can be ignored since $b$ should be placed between $\pi^+$ and $\pi^-$.
Therefore, the recent RL methods~\citep{schulman2015trust,tsurumine2019deep,kobayashi2020proximal} focus on the constraint between $\pi$ and $b$, which contributes to the improvement of learning performance.
In reality, however, $\pi$ is not equivalent with $b$ especially during the experience replay.

Another direction for solving this issue is to add regularization to the value function so as to force $\pi^+$ and $\pi^-$ to diverge from each other.
If this regularization works correctly, it should increase the probability that $\pi$ is placed between $\pi^+$ and $\pi^-$, thereby stabilizing policy learning.
However, it may make the optimality deterministic too much, and the risk of getting one of the local solutions would be increased.

In the field of metric learning using triplet loss, the solutions to this issue have been proposed.
For example, one of them~\citep{cheng2016person} can increase the possibility of making $\pi$ reach $\pi^+$ by adding a minimization problem of the divergence between $\pi$ and $\pi^+$.
The weight of this additional term is however difficult to be designed.

As mentioned above, the fact that policy learning is expressed as the triplet-loss-based optimization problem suggests new directions for stabilizing learning.
Using the above directions as stepping stones, it is believed that RL can be evolved so as to learn the optimal policy stably without affected by random seeds.

The remaining challenge in deriving the update law is biased but consistent approximation to eliminate the uncomputable $p_V$ with $C$ the unknown maximum value.
Indeed, RKL-RL is consistent with the traditional RL thanks to this approximation, but its properties with the exact gradients should be investigated.
To this end, it would be required that a new problem statement in which $C$ is clearly given and/or a new definition of optimality that does not use such an unknown variable.
The simplest method is to use a task with a given upper bound on the reward, in which case we can evaluate the effect of bias by comparing the methods with the exact gradients to RKL-RL and FKL-RL proposed in this paper.

\section{Conclusion}
\label{sec:conclusion}

This paper addressed a new interpretation of of the traditional optimization method in RL as reverse KL divergence optimization, and derived a new optimization method using forward KL divergence, so-called FKL-RL.
Specifically, the way of updating the value function was derived as the minimization problem of KL divergence between the probabilities of optimality with different conditions.
In addition, with optimal/non-optimal policies inferred from Bayesian theorem, the policy improvement was also derived as the combination of minimizing KL divergence for the optimal policy and maximizing KL divergence for the non-optimal policy.
If reverse KL divergence is employed for these optimization problems, their update laws are consistent with the traditional RL method; if forward KL divergence is employed, one can obtain a new optimization method, FKL-RL.
One remarkable difference was found in FKL-RL that the similar but different learning laws from the traditional RL yield biological optimism.
By heuristically converting the original uncertainty parameter $\tau$ into the optimism parameter $\eta$, the optimism in FKL-RL can be adaptively designed.
The effects of this optimism was investigated through learning tendencies on numerical simulations using Pybullet.
As a result, the optimism gained by FKL-RL certainly improved the capability to find the global solution by keeping exploration toward probably better situations, although too much optimism caused the increase of learning variance.
In more practical robot simulations, FKL-RL was able to obtain performance better than PPO algorithm.

The new derivation of RL, so to speak \textit{RL as divergence optimization}, makes us expect not only FKL-RL but also methods with different properties derived from other divergences.
In the future, this concept will be generalized for arbitrary divergences and investigated towards automatic selection of the task-specific divergence.
In addition, a new optimization problem, which solves the problem caused by triplet loss as mentioned in the above discussion, will be derived for more stable learning.

\appendix
\section{Implementation details}
\label{app:algorithm}

\begin{algorithm*}[tb]
    \caption{Implemented algorithm}
    \label{alg:algorithm}
    \begin{algorithmic}[1]
        \State{Given environment as $p_0(s)$, $p_e(s^\prime \mid s, a)$, and $r(s, a)$}
        \State{Given the hyperparameters from Table~\ref{tab:parameter}}
        \State{Initialize the value function $V(s)$ with $\theta$ and the policy $\pi(a \mid s)$ with $\phi$ as the main networks}
        \State{$\bar{\theta} \gets \theta$ and $\bar{\phi} \gets \phi$ for the target networks, $\bar{V}(s)$ and $b(a \mid s)$, respectively}
        \State{Initialize $\Delta_\mathrm{max} = \Delta = \epsilon^{-1}$}
        \State{Initialize the replay buffer $\mathcal{D} \gets \emptyset$}
        \While{True}
            \State{Reset eligibility traces $e \gets 0$}
            \State{Get the initial state $s \sim p_0(s)$}
            \While{Until the end of the episode}
                \State{Get the action for interaction $a \sim b(a \mid s)$}
                \State{Get the next state $s^\prime \sim p_e(s^\prime \mid s, a)$}
                \State{Get the reward $r \gets r(s, a)$}
                \State{Compute the TD error $\delta \gets r + \gamma \bar{V}(s^\prime) - V(s)$}
                \State{Update the uncertainty $\tau$ by eqs.~\eqref{eq:err_max}--\eqref{eq:tau_adapt2}}
                \State{Compute the gradients $g_{\theta, \phi}^{R, L}$ by eq.~\eqref{eq:grad_rkl} or~\eqref{eq:grad_fkl} with one sample for Monte Carlo approximation}
                \State{Accumulate $g_{\theta, \phi}^{R, L} / \delta$ as the alternative eligibility traces $e$}
                \State{Update $\theta$ and $\phi$ by $\delta e$}
                \State{Update $\bar{\theta}$ and $\bar{\phi}$ by t-soft update}
                \State{Store the data $\mathcal{D} \gets \mathcal{D} \cup (s, a, s^\prime, r, |\delta|, b(a \mid s))$}
                \State{Delete the oldest data if $|\mathcal{D}| > N_c$}
                \State{$s \gets s^\prime$}
            \EndWhile
            \For{$i \gets 1:N_e$}
                \State{Sample the buffer $\mathcal{B}_i \sim \mathcal{D}$ by PER with eq.~\eqref{eq:per_weight_origin} or~\eqref{eq:per_weight_fkl}}
                \State{Compute the TD error $\delta \gets r + \gamma \bar{V}(s^\prime) - V(s)$}
                \State{Update the uncertainty $\tau$ by eqs.~\eqref{eq:err_max}--\eqref{eq:tau_adapt2}}
                \State{Compute the gradients $g_{\theta, \phi}^{R, L}$ by eq.~\eqref{eq:grad_rkl} or~\eqref{eq:grad_fkl} with $N_b$ samples for Monte Carlo approximation}
                \State{Update $\theta$ and $\phi$ by $g_{\theta, \phi}^{R, L}$}
                \State{Update $\bar{\theta}$ and $\bar{\phi}$ by t-soft update}
                \State{Update $|\delta|$ in $\mathcal{D}$}
            \EndFor
        \EndWhile
    \end{algorithmic}
\end{algorithm*}

A pseudo code of the implemented algorithm is described in Alg.~\ref{alg:algorithm}.
The target networks are used for sampling actions, $b$, and computing the target signal for the value function, $r + \gamma V(s^\prime)$.
They follow the main networks by t-soft update~\citep{kobayashi2021t} immediately after each of the following updates.
Note that in the post-training test, actions are sampled from $\pi$ instead of $b$, although $b \simeq \pi$ is mostly satisfied.

Learning is divided into two parts: during the episode and after the episode.
During the episode, the value function and the policy are updated with the eligibility traces method~\citep{kobayashi2020adaptive} along with the state-action trajectory.
Note that, although FKL-RL is off-policy learning, the eligibility traces method can be applied when using such a trajectory.
After the episode, $N_e$ batches with $N_b$ size for each are sampled from the stored replay buffer by PER~\citep{schaul2015prioritized}, and the mean of the gradients in each batch is computed for updating.
Note that, since RKL-RL is on-policy learning, the likelihood at the time of action samples, $b(a \mid s)$, is also stored in the replay buffer to compute the density ratio $\rho$.

As a remark, PPO used in the second experiment actually refers to RKL-RL with a clipping-based policy regularization only, not its original implementation~\citep{schulman2017proximal}.
Specifically, when computing the density ratio $\rho$ in RKL-RL, it is clipped if it is over a threshold.
This implementation is equivalent to the implementation in \citet{kobayashi2020proximal}, although the different hyperparameters are used.
Note that, even with the different hyperparameters, the implemented PPO can accomplish the other RL benchmarks.

\section{Configurations of Minitaur}
\label{app:minitaur}

\textit{MinitaurBulletEnv-v0} abbreviated as Minitaur is for making a quadruped robot walk as fast as possible.
Each leg of the quadruped robot has two actuators with velocity limitation (not $\infty$ in reality), and therefore, the action space is eight-dimensional.
In general, the kinematics of the controlled robot is well-known, but the dynamics of the robot (even only actuators) would not be identified precisely.
Hence, PD control is employed instead of inverse dynamics.
Although the default control period is 100~Hz, due to the limitations of observation, command communication, and/or computation of DRL in real systems, it is reduced to 25~Hz by repeating the same action four times.
This task has 28-dimensional state space, and in reality, they are observed with noise.
In real robotic systems, hard reset is time-consuming, so it should be omitted.
The simulation model is not changed in each episode due to uniqueness of real robot.
In summary, the arguments in this task are described in Table~\ref{tab:minitaur}.
The omitted arguments are with the default values.
Note that the weights of the rewards are also adjusted in order to satisfy realistic demands and scale the rewards with other tasks to some extent.

\begin{table}[tb]
    \caption{Arguments for \textit{MinitaurBulletEnv-v0}}
    \label{tab:minitaur}
    \centering
    \begin{tabular}{ccc}
        \hline\hline
        Name & Default & Modified
        \\ \hline
        \texttt{motor\_velocity\_limit} & $\infty$ & 1000
        \\
        \texttt{pd\_control\_enabled} & False & True
        \\
        \texttt{accurate\_motor\_model\_enabled} & True & False
        \\
        \texttt{action\_repeat} & 1 & 4
        \\
        \texttt{observation\_noise\_stdev} & 0 & 0.0001
        \\
        \texttt{hard\_reset} & True & False
        \\
        \texttt{env\_randomizer} & Uniform & None
        \\
        \texttt{distance\_weight} & 1 & 200
        \\
        \texttt{energy\_weight} & 0.005 & 1
        \\
        \texttt{shake\_weight} & 0 & 1
        \\
        \texttt{drift\_weight} & 0 & 1
        \\
        \hline\hline
    \end{tabular}
\end{table}

\bibliographystyle{elsarticle-harv}
\bibliography{biblio}

\begin{thebibliography}{53}
\expandafter\ifx\csname natexlab\endcsname\relax\def\natexlab#1{#1}\fi
\providecommand{\url}[1]{\texttt{#1}}
\providecommand{\href}[2]{#2}
\providecommand{\path}[1]{#1}
\providecommand{\DOIprefix}{doi:}
\providecommand{\ArXivprefix}{arXiv:}
\providecommand{\URLprefix}{URL: }
\providecommand{\Pubmedprefix}{pmid:}
\providecommand{\doi}[1]{\href{http://dx.doi.org/#1}{\path{#1}}}
\providecommand{\Pubmed}[1]{\href{pmid:#1}{\path{#1}}}
\providecommand{\bibinfo}[2]{#2}
\ifx\xfnm\relax \def\xfnm[#1]{\unskip,\space#1}\fi
\bibitem[{Andrychowicz et~al.(2017)Andrychowicz, Wolski, Ray, Schneider, Fong,
  Welinder, McGrew, Tobin, Abbeel and Zaremba}]{andrychowicz2017hindsight}
\bibinfo{author}{Andrychowicz, M.}, \bibinfo{author}{Wolski, F.},
  \bibinfo{author}{Ray, A.}, \bibinfo{author}{Schneider, J.},
  \bibinfo{author}{Fong, R.}, \bibinfo{author}{Welinder, P.},
  \bibinfo{author}{McGrew, B.}, \bibinfo{author}{Tobin, J.},
  \bibinfo{author}{Abbeel, O.P.}, \bibinfo{author}{Zaremba, W.},
  \bibinfo{year}{2017}.
\newblock \bibinfo{title}{Hindsight experience replay}, in:
  \bibinfo{booktitle}{Advances in Neural Information Processing Systems}, pp.
  \bibinfo{pages}{5048--5058}.
\bibitem[{Ba et~al.(2016)Ba, Kiros and Hinton}]{ba2016layer}
\bibinfo{author}{Ba, J.L.}, \bibinfo{author}{Kiros, J.R.},
  \bibinfo{author}{Hinton, G.E.}, \bibinfo{year}{2016}.
\newblock \bibinfo{title}{Layer normalization}.
\newblock \bibinfo{journal}{arXiv preprint arXiv:1607.06450} .
\bibitem[{Brockman et~al.(2016)Brockman, Cheung, Pettersson, Schneider,
  Schulman, Tang and Zaremba}]{brockman2016openai}
\bibinfo{author}{Brockman, G.}, \bibinfo{author}{Cheung, V.},
  \bibinfo{author}{Pettersson, L.}, \bibinfo{author}{Schneider, J.},
  \bibinfo{author}{Schulman, J.}, \bibinfo{author}{Tang, J.},
  \bibinfo{author}{Zaremba, W.}, \bibinfo{year}{2016}.
\newblock \bibinfo{title}{Openai gym}.
\newblock \bibinfo{journal}{arXiv preprint arXiv:1606.01540} .
\bibitem[{Chechik et~al.(2010)Chechik, Sharma, Shalit and
  Bengio}]{chechik2010large}
\bibinfo{author}{Chechik, G.}, \bibinfo{author}{Sharma, V.},
  \bibinfo{author}{Shalit, U.}, \bibinfo{author}{Bengio, S.},
  \bibinfo{year}{2010}.
\newblock \bibinfo{title}{Large scale online learning of image similarity
  through ranking}.
\newblock \bibinfo{journal}{Journal of Machine Learning Research}
  \bibinfo{volume}{11}, \bibinfo{pages}{1109--1135}.
\bibitem[{Cheng et~al.(2016)Cheng, Gong, Zhou, Wang and
  Zheng}]{cheng2016person}
\bibinfo{author}{Cheng, D.}, \bibinfo{author}{Gong, Y.}, \bibinfo{author}{Zhou,
  S.}, \bibinfo{author}{Wang, J.}, \bibinfo{author}{Zheng, N.},
  \bibinfo{year}{2016}.
\newblock \bibinfo{title}{Person re-identification by multi-channel parts-based
  cnn with improved triplet loss function}, in: \bibinfo{booktitle}{Proceedings
  of the iEEE conference on computer vision and pattern recognition}, pp.
  \bibinfo{pages}{1335--1344}.
\bibitem[{Chua et~al.(2018)Chua, Calandra, McAllister and
  Levine}]{chua2018deep}
\bibinfo{author}{Chua, K.}, \bibinfo{author}{Calandra, R.},
  \bibinfo{author}{McAllister, R.}, \bibinfo{author}{Levine, S.},
  \bibinfo{year}{2018}.
\newblock \bibinfo{title}{Deep reinforcement learning in a handful of trials
  using probabilistic dynamics models}, in: \bibinfo{booktitle}{Advances in
  Neural Information Processing Systems}, pp. \bibinfo{pages}{4754--4765}.
\bibitem[{Clavera et~al.(2020)Clavera, Fu and Abbeel}]{clavera2019model}
\bibinfo{author}{Clavera, I.}, \bibinfo{author}{Fu, Y.},
  \bibinfo{author}{Abbeel, P.}, \bibinfo{year}{2020}.
\newblock \bibinfo{title}{Model-augmented actor-critic: Backpropagating through
  paths}, in: \bibinfo{booktitle}{International Conference on Learning
  Representations}.
\bibitem[{Colas et~al.(2018)Colas, Sigaud and Oudeyer}]{colas2018many}
\bibinfo{author}{Colas, C.}, \bibinfo{author}{Sigaud, O.},
  \bibinfo{author}{Oudeyer, P.Y.}, \bibinfo{year}{2018}.
\newblock \bibinfo{title}{How many random seeds? statistical power analysis in
  deep reinforcement learning experiments}.
\newblock \bibinfo{journal}{arXiv preprint arXiv:1806.08295} .
\bibitem[{Coumans and Bai(2016)}]{coumans2016pybullet}
\bibinfo{author}{Coumans, E.}, \bibinfo{author}{Bai, Y.}, \bibinfo{year}{2016}.
\newblock \bibinfo{title}{Pybullet, a python module for physics simulation for
  games, robotics and machine learning}.
\newblock \bibinfo{journal}{GitHub repository} .
\bibitem[{Curi et~al.(2020)Curi, Berkenkamp and Krause}]{curi2020efficient}
\bibinfo{author}{Curi, S.}, \bibinfo{author}{Berkenkamp, F.},
  \bibinfo{author}{Krause, A.}, \bibinfo{year}{2020}.
\newblock \bibinfo{title}{Efficient model-based reinforcement learning through
  optimistic policy search and planning}.
\newblock \bibinfo{journal}{Advances in Neural Information Processing Systems}
  \bibinfo{volume}{33}.
\bibitem[{Daniel et~al.(2016)Daniel, Neumann, Kroemer, Peters
  et~al.}]{daniel2016hierarchical}
\bibinfo{author}{Daniel, C.}, \bibinfo{author}{Neumann, G.},
  \bibinfo{author}{Kroemer, O.}, \bibinfo{author}{Peters, J.}, et~al.,
  \bibinfo{year}{2016}.
\newblock \bibinfo{title}{Hierarchical relative entropy policy search}.
\newblock \bibinfo{journal}{Journal of Machine Learning Research}
  \bibinfo{volume}{17}, \bibinfo{pages}{1--50}.
\bibitem[{Elfwing et~al.(2018)Elfwing, Uchibe and Doya}]{elfwing2018sigmoid}
\bibinfo{author}{Elfwing, S.}, \bibinfo{author}{Uchibe, E.},
  \bibinfo{author}{Doya, K.}, \bibinfo{year}{2018}.
\newblock \bibinfo{title}{Sigmoid-weighted linear units for neural network
  function approximation in reinforcement learning}.
\newblock \bibinfo{journal}{Neural Networks} \bibinfo{volume}{107},
  \bibinfo{pages}{3--11}.
\bibitem[{Fujimoto et~al.(2018)Fujimoto, Hoof and
  Meger}]{fujimoto2018addressing}
\bibinfo{author}{Fujimoto, S.}, \bibinfo{author}{Hoof, H.},
  \bibinfo{author}{Meger, D.}, \bibinfo{year}{2018}.
\newblock \bibinfo{title}{Addressing function approximation error in
  actor-critic methods}, in: \bibinfo{booktitle}{International Conference on
  Machine Learning}, \bibinfo{organization}{PMLR}. pp.
  \bibinfo{pages}{1587--1596}.
\bibitem[{Greensmith et~al.(2004)Greensmith, Bartlett and
  Baxter}]{greensmith2004variance}
\bibinfo{author}{Greensmith, E.}, \bibinfo{author}{Bartlett, P.L.},
  \bibinfo{author}{Baxter, J.}, \bibinfo{year}{2004}.
\newblock \bibinfo{title}{Variance reduction techniques for gradient estimates
  in reinforcement learning.}
\newblock \bibinfo{journal}{Journal of Machine Learning Research}
  \bibinfo{volume}{5}.
\bibitem[{Haarnoja et~al.(2018)Haarnoja, Zhou, Abbeel and
  Levine}]{haarnoja2018soft}
\bibinfo{author}{Haarnoja, T.}, \bibinfo{author}{Zhou, A.},
  \bibinfo{author}{Abbeel, P.}, \bibinfo{author}{Levine, S.},
  \bibinfo{year}{2018}.
\newblock \bibinfo{title}{Soft actor-critic: Off-policy maximum entropy deep
  reinforcement learning with a stochastic actor}.
\newblock \bibinfo{journal}{arXiv preprint arXiv:1801.01290} .
\bibitem[{Henderson et~al.(2018)Henderson, Islam, Bachman, Pineau, Precup and
  Meger}]{henderson2018deep}
\bibinfo{author}{Henderson, P.}, \bibinfo{author}{Islam, R.},
  \bibinfo{author}{Bachman, P.}, \bibinfo{author}{Pineau, J.},
  \bibinfo{author}{Precup, D.}, \bibinfo{author}{Meger, D.},
  \bibinfo{year}{2018}.
\newblock \bibinfo{title}{Deep reinforcement learning that matters}, in:
  \bibinfo{booktitle}{Proceedings of the AAAI Conference on Artificial
  Intelligence}.
\bibitem[{Ilboudo et~al.(2020)Ilboudo, Kobayashi and
  Sugimoto}]{ilboudo2020robust}
\bibinfo{author}{Ilboudo, W.E.L.}, \bibinfo{author}{Kobayashi, T.},
  \bibinfo{author}{Sugimoto, K.}, \bibinfo{year}{2020}.
\newblock \bibinfo{title}{Robust stochastic gradient descent with student-t
  distribution based first-order momentum}.
\newblock \bibinfo{journal}{IEEE Transactions on Neural Networks and Learning
  Systems} , \bibinfo{pages}{1--14}.
\bibitem[{Ke et~al.(2019)Ke, Barnes, Sun, Lee, Choudhury and
  Srinivasa}]{ke2019imitation}
\bibinfo{author}{Ke, L.}, \bibinfo{author}{Barnes, M.}, \bibinfo{author}{Sun,
  W.}, \bibinfo{author}{Lee, G.}, \bibinfo{author}{Choudhury, S.},
  \bibinfo{author}{Srinivasa, S.}, \bibinfo{year}{2019}.
\newblock \bibinfo{title}{Imitation learning as $ f $-divergence minimization}.
\newblock \bibinfo{journal}{arXiv preprint arXiv:1905.12888} .
\bibitem[{Kobayashi(2020a)}]{kobayashi2020adaptive}
\bibinfo{author}{Kobayashi, T.}, \bibinfo{year}{2020}a.
\newblock \bibinfo{title}{Adaptive and multiple time-scale eligibility traces
  for online deep reinforcement learning}.
\newblock \bibinfo{journal}{arXiv preprint arXiv:2008.10040} .
\bibitem[{Kobayashi(2020b)}]{kobayashi2020proximal}
\bibinfo{author}{Kobayashi, T.}, \bibinfo{year}{2020}b.
\newblock \bibinfo{title}{Proximal policy optimization with relative pearson
  divergence}.
\newblock \bibinfo{journal}{arXiv preprint arXiv:2010.03290} .
\bibitem[{Kobayashi and Ilboudo(2021)}]{kobayashi2021t}
\bibinfo{author}{Kobayashi, T.}, \bibinfo{author}{Ilboudo, W.E.L.},
  \bibinfo{year}{2021}.
\newblock \bibinfo{title}{t-soft update of target network for deep
  reinforcement learning}.
\newblock \bibinfo{journal}{Neural Networks} \bibinfo{volume}{136},
  \bibinfo{pages}{63--71}.
\bibitem[{Kormushev et~al.(2010)Kormushev, Calinon and
  Caldwell}]{kormushev2010robot}
\bibinfo{author}{Kormushev, P.}, \bibinfo{author}{Calinon, S.},
  \bibinfo{author}{Caldwell, D.G.}, \bibinfo{year}{2010}.
\newblock \bibinfo{title}{Robot motor skill coordination with em-based
  reinforcement learning}, in: \bibinfo{booktitle}{IEEE/RSJ international
  conference on intelligent robots and systems}, \bibinfo{organization}{IEEE}.
  pp. \bibinfo{pages}{3232--3237}.
\bibitem[{Kullback(1997)}]{kullback1997information}
\bibinfo{author}{Kullback, S.}, \bibinfo{year}{1997}.
\newblock \bibinfo{title}{Information theory and statistics}.
\newblock \bibinfo{publisher}{Courier Corporation}.
\bibitem[{LeCun et~al.(2015)LeCun, Bengio and Hinton}]{lecun2015deep}
\bibinfo{author}{LeCun, Y.}, \bibinfo{author}{Bengio, Y.},
  \bibinfo{author}{Hinton, G.}, \bibinfo{year}{2015}.
\newblock \bibinfo{title}{Deep learning}.
\newblock \bibinfo{journal}{nature} \bibinfo{volume}{521},
  \bibinfo{pages}{436}.
\bibitem[{Lee et~al.(2020)Lee, Nagabandi, Abbeel and
  Levine}]{lee2020stochastic}
\bibinfo{author}{Lee, A.}, \bibinfo{author}{Nagabandi, A.},
  \bibinfo{author}{Abbeel, P.}, \bibinfo{author}{Levine, S.},
  \bibinfo{year}{2020}.
\newblock \bibinfo{title}{Stochastic latent actor-critic: Deep reinforcement
  learning with a latent variable model}.
\newblock \bibinfo{journal}{Advances in Neural Information Processing Systems}
  \bibinfo{volume}{33}.
\bibitem[{Lefebvre et~al.(2017)Lefebvre, Lebreton, Meyniel, Bourgeois-Gironde
  and Palminteri}]{lefebvre2017behavioural}
\bibinfo{author}{Lefebvre, G.}, \bibinfo{author}{Lebreton, M.},
  \bibinfo{author}{Meyniel, F.}, \bibinfo{author}{Bourgeois-Gironde, S.},
  \bibinfo{author}{Palminteri, S.}, \bibinfo{year}{2017}.
\newblock \bibinfo{title}{Behavioural and neural characterization of optimistic
  reinforcement learning}.
\newblock \bibinfo{journal}{Nature Human Behaviour} \bibinfo{volume}{1},
  \bibinfo{pages}{1--9}.
\bibitem[{Levine(2018)}]{levine2018reinforcement}
\bibinfo{author}{Levine, S.}, \bibinfo{year}{2018}.
\newblock \bibinfo{title}{Reinforcement learning and control as probabilistic
  inference: Tutorial and review}.
\newblock \bibinfo{journal}{arXiv preprint arXiv:1805.00909} .
\bibitem[{Levine et~al.(2018)Levine, Pastor, Krizhevsky, Ibarz and
  Quillen}]{levine2018learning}
\bibinfo{author}{Levine, S.}, \bibinfo{author}{Pastor, P.},
  \bibinfo{author}{Krizhevsky, A.}, \bibinfo{author}{Ibarz, J.},
  \bibinfo{author}{Quillen, D.}, \bibinfo{year}{2018}.
\newblock \bibinfo{title}{Learning hand-eye coordination for robotic grasping
  with deep learning and large-scale data collection}.
\newblock \bibinfo{journal}{The International Journal of Robotics Research}
  \bibinfo{volume}{37}, \bibinfo{pages}{421--436}.
\bibitem[{Lin(1992)}]{lin1992self}
\bibinfo{author}{Lin, L.J.}, \bibinfo{year}{1992}.
\newblock \bibinfo{title}{Self-improving reactive agents based on reinforcement
  learning, planning and teaching}.
\newblock \bibinfo{journal}{Machine learning} \bibinfo{volume}{8},
  \bibinfo{pages}{293--321}.
\bibitem[{Machado et~al.(2015)Machado, Srinivasan and
  Bowling}]{machado2015domain}
\bibinfo{author}{Machado, M.C.}, \bibinfo{author}{Srinivasan, S.},
  \bibinfo{author}{Bowling, M.H.}, \bibinfo{year}{2015}.
\newblock \bibinfo{title}{Domain-independent optimistic initialization for
  reinforcement learning}, in: \bibinfo{booktitle}{AAAI Workshop: Learning for
  General Competency in Video Games}.
\bibitem[{Modares et~al.(2015)Modares, Ranatunga, Lewis and
  Popa}]{modares2015optimized}
\bibinfo{author}{Modares, H.}, \bibinfo{author}{Ranatunga, I.},
  \bibinfo{author}{Lewis, F.L.}, \bibinfo{author}{Popa, D.O.},
  \bibinfo{year}{2015}.
\newblock \bibinfo{title}{Optimized assistive human--robot interaction using
  reinforcement learning}.
\newblock \bibinfo{journal}{IEEE transactions on cybernetics}
  \bibinfo{volume}{46}, \bibinfo{pages}{655--667}.
\bibitem[{Munos et~al.(2016)Munos, Stepleton, Harutyunyan and
  Bellemare}]{munos2016safe}
\bibinfo{author}{Munos, R.}, \bibinfo{author}{Stepleton, T.},
  \bibinfo{author}{Harutyunyan, A.}, \bibinfo{author}{Bellemare, M.G.},
  \bibinfo{year}{2016}.
\newblock \bibinfo{title}{Safe and efficient off-policy reinforcement
  learning}, in: \bibinfo{booktitle}{International Conference on Neural
  Information Processing Systems}, pp. \bibinfo{pages}{1054--1062}.
\bibitem[{Oh et~al.(2018)Oh, Guo, Singh and Lee}]{oh2018self}
\bibinfo{author}{Oh, J.}, \bibinfo{author}{Guo, Y.}, \bibinfo{author}{Singh,
  S.}, \bibinfo{author}{Lee, H.}, \bibinfo{year}{2018}.
\newblock \bibinfo{title}{Self-imitation learning}, in:
  \bibinfo{booktitle}{International Conference on Machine Learning},
  \bibinfo{organization}{PMLR}. pp. \bibinfo{pages}{3878--3887}.
\bibitem[{Okada and Taniguchi(2020)}]{okada2020variational}
\bibinfo{author}{Okada, M.}, \bibinfo{author}{Taniguchi, T.},
  \bibinfo{year}{2020}.
\newblock \bibinfo{title}{Variational inference mpc for bayesian model-based
  reinforcement learning}, in: \bibinfo{booktitle}{Conference on Robot
  Learning}, \bibinfo{organization}{PMLR}. pp. \bibinfo{pages}{258--272}.
\bibitem[{Parisi et~al.(2019)Parisi, Tangkaratt, Peters and
  Khan}]{parisi2019td}
\bibinfo{author}{Parisi, S.}, \bibinfo{author}{Tangkaratt, V.},
  \bibinfo{author}{Peters, J.}, \bibinfo{author}{Khan, M.E.},
  \bibinfo{year}{2019}.
\newblock \bibinfo{title}{Td-regularized actor-critic methods}.
\newblock \bibinfo{journal}{Machine Learning} , \bibinfo{pages}{1--35}.
\bibitem[{Paszke et~al.(2017)Paszke, Gross, Chintala, Chanan, Yang, DeVito,
  Lin, Desmaison, Antiga and Lerer}]{paszke2017automatic}
\bibinfo{author}{Paszke, A.}, \bibinfo{author}{Gross, S.},
  \bibinfo{author}{Chintala, S.}, \bibinfo{author}{Chanan, G.},
  \bibinfo{author}{Yang, E.}, \bibinfo{author}{DeVito, Z.},
  \bibinfo{author}{Lin, Z.}, \bibinfo{author}{Desmaison, A.},
  \bibinfo{author}{Antiga, L.}, \bibinfo{author}{Lerer, A.},
  \bibinfo{year}{2017}.
\newblock \bibinfo{title}{Automatic differentiation in pytorch}, in:
  \bibinfo{booktitle}{Advances in Neural Information Processing Systems
  Workshop}.
\bibitem[{Peng et~al.(2017)Peng, Berseth, Yin and Van
  De~Panne}]{peng2017deeploco}
\bibinfo{author}{Peng, X.B.}, \bibinfo{author}{Berseth, G.},
  \bibinfo{author}{Yin, K.}, \bibinfo{author}{Van De~Panne, M.},
  \bibinfo{year}{2017}.
\newblock \bibinfo{title}{Deeploco: Dynamic locomotion skills using
  hierarchical deep reinforcement learning}.
\newblock \bibinfo{journal}{ACM Transactions on Graphics} \bibinfo{volume}{36},
  \bibinfo{pages}{1--13}.
\bibitem[{Rashid et~al.(2020)Rashid, Peng, Boehmer and
  Whiteson}]{rashid2019optimistic}
\bibinfo{author}{Rashid, T.}, \bibinfo{author}{Peng, B.},
  \bibinfo{author}{Boehmer, W.}, \bibinfo{author}{Whiteson, S.},
  \bibinfo{year}{2020}.
\newblock \bibinfo{title}{Optimistic exploration even with a pessimistic
  initialisation}, in: \bibinfo{booktitle}{International Conference on Learning
  Representations}.
\bibitem[{Sasaki and Matsubara(2019)}]{sasaki2019multimodal}
\bibinfo{author}{Sasaki, H.}, \bibinfo{author}{Matsubara, T.},
  \bibinfo{year}{2019}.
\newblock \bibinfo{title}{Multimodal policy search using overlapping mixtures
  of sparse gaussian process prior}, in: \bibinfo{booktitle}{International
  Conference on Robotics and Automation}, \bibinfo{organization}{IEEE}. pp.
  \bibinfo{pages}{2433--2439}.
\bibitem[{Schaul et~al.(2015)Schaul, Quan, Antonoglou and
  Silver}]{schaul2015prioritized}
\bibinfo{author}{Schaul, T.}, \bibinfo{author}{Quan, J.},
  \bibinfo{author}{Antonoglou, I.}, \bibinfo{author}{Silver, D.},
  \bibinfo{year}{2015}.
\newblock \bibinfo{title}{Prioritized experience replay}.
\newblock \bibinfo{journal}{arXiv preprint arXiv:1511.05952} .
\bibitem[{Schulman et~al.(2015)Schulman, Levine, Abbeel, Jordan and
  Moritz}]{schulman2015trust}
\bibinfo{author}{Schulman, J.}, \bibinfo{author}{Levine, S.},
  \bibinfo{author}{Abbeel, P.}, \bibinfo{author}{Jordan, M.},
  \bibinfo{author}{Moritz, P.}, \bibinfo{year}{2015}.
\newblock \bibinfo{title}{Trust region policy optimization}, in:
  \bibinfo{booktitle}{International conference on machine learning},
  \bibinfo{organization}{PMLR}. pp. \bibinfo{pages}{1889--1897}.
\bibitem[{Schulman et~al.(2016)Schulman, Moritz, Levine, Jordan and
  Abbeel}]{schulman2015high}
\bibinfo{author}{Schulman, J.}, \bibinfo{author}{Moritz, P.},
  \bibinfo{author}{Levine, S.}, \bibinfo{author}{Jordan, M.},
  \bibinfo{author}{Abbeel, P.}, \bibinfo{year}{2016}.
\newblock \bibinfo{title}{High-dimensional continuous control using generalized
  advantage estimation}, in: \bibinfo{booktitle}{International Conference on
  Learning Representations}.
\bibitem[{Schulman et~al.(2017)Schulman, Wolski, Dhariwal, Radford and
  Klimov}]{schulman2017proximal}
\bibinfo{author}{Schulman, J.}, \bibinfo{author}{Wolski, F.},
  \bibinfo{author}{Dhariwal, P.}, \bibinfo{author}{Radford, A.},
  \bibinfo{author}{Klimov, O.}, \bibinfo{year}{2017}.
\newblock \bibinfo{title}{Proximal policy optimization algorithms}.
\newblock \bibinfo{journal}{arXiv preprint arXiv:1707.06347} .
\bibitem[{Schultz and Joachims(2003)}]{schultz2003learning}
\bibinfo{author}{Schultz, M.}, \bibinfo{author}{Joachims, T.},
  \bibinfo{year}{2003}.
\newblock \bibinfo{title}{Learning a distance metric from relative
  comparisons}.
\newblock \bibinfo{journal}{Advances in Neural Information Processing Systems}
  \bibinfo{volume}{16}, \bibinfo{pages}{41--48}.
\bibitem[{van Seijen(2016)}]{van2016effective}
\bibinfo{author}{van Seijen, H.}, \bibinfo{year}{2016}.
\newblock \bibinfo{title}{Effective multi-step temporal-difference learning for
  non-linear function approximation}.
\newblock \bibinfo{journal}{arXiv preprint arXiv:1608.05151} .
\bibitem[{Shi et~al.(2019)Shi, Song and Wu}]{shi2019soft}
\bibinfo{author}{Shi, W.}, \bibinfo{author}{Song, S.}, \bibinfo{author}{Wu,
  C.}, \bibinfo{year}{2019}.
\newblock \bibinfo{title}{Soft policy gradient method for maximum entropy deep
  reinforcement learning}, in: \bibinfo{booktitle}{International Joint
  Conference on Artificial Intelligence}, pp. \bibinfo{pages}{3425--3431}.
\bibitem[{Sunehag and Hutter(2015)}]{sunehag2015rationality}
\bibinfo{author}{Sunehag, P.}, \bibinfo{author}{Hutter, M.},
  \bibinfo{year}{2015}.
\newblock \bibinfo{title}{Rationality, optimism and guarantees in general
  reinforcement learning}.
\newblock \bibinfo{journal}{The Journal of Machine Learning Research}
  \bibinfo{volume}{16}, \bibinfo{pages}{1345--1390}.
\bibitem[{Sutton and Barto(2018)}]{sutton2018reinforcement}
\bibinfo{author}{Sutton, R.S.}, \bibinfo{author}{Barto, A.G.},
  \bibinfo{year}{2018}.
\newblock \bibinfo{title}{Reinforcement learning: An introduction}.
\newblock \bibinfo{publisher}{MIT press}.
\bibitem[{Tokdar and Kass(2010)}]{tokdar2010importance}
\bibinfo{author}{Tokdar, S.T.}, \bibinfo{author}{Kass, R.E.},
  \bibinfo{year}{2010}.
\newblock \bibinfo{title}{Importance sampling: a review}.
\newblock \bibinfo{journal}{Wiley Interdisciplinary Reviews: Computational
  Statistics} \bibinfo{volume}{2}, \bibinfo{pages}{54--60}.
\bibitem[{Tsurumine et~al.(2019)Tsurumine, Cui, Uchibe and
  Matsubara}]{tsurumine2019deep}
\bibinfo{author}{Tsurumine, Y.}, \bibinfo{author}{Cui, Y.},
  \bibinfo{author}{Uchibe, E.}, \bibinfo{author}{Matsubara, T.},
  \bibinfo{year}{2019}.
\newblock \bibinfo{title}{Deep reinforcement learning with smooth policy
  update: Application to robotic cloth manipulation}.
\newblock \bibinfo{journal}{Robotics and Autonomous Systems}
  \bibinfo{volume}{112}, \bibinfo{pages}{72--83}.
\bibitem[{Uchibe and Doya(2020)}]{uchibe2020imitation}
\bibinfo{author}{Uchibe, E.}, \bibinfo{author}{Doya, K.}, \bibinfo{year}{2020}.
\newblock \bibinfo{title}{Imitation learning based on entropy-regularized
  forward and inverse reinforcement learning}.
\newblock \bibinfo{journal}{arXiv preprint arXiv:2008.07284} .
\bibitem[{Vuong and Tran(2019)}]{vuong2019uncertainty}
\bibinfo{author}{Vuong, T.L.}, \bibinfo{author}{Tran, K.},
  \bibinfo{year}{2019}.
\newblock \bibinfo{title}{Uncertainty-aware model-based policy optimization}.
\newblock \bibinfo{journal}{arXiv preprint arXiv:1906.10717} .
\bibitem[{Zaheer et~al.(2018)Zaheer, Reddi, Sachan, Kale and
  Kumar}]{zaheer2018adaptive}
\bibinfo{author}{Zaheer, M.}, \bibinfo{author}{Reddi, S.J.},
  \bibinfo{author}{Sachan, D.}, \bibinfo{author}{Kale, S.},
  \bibinfo{author}{Kumar, S.}, \bibinfo{year}{2018}.
\newblock \bibinfo{title}{Adaptive methods for nonconvex optimization}, in:
  \bibinfo{booktitle}{International Conference on Neural Information Processing
  Systems}, pp. \bibinfo{pages}{9815--9825}.

\end{thebibliography}

\end{document}